\documentclass[11pt,english]{article}
\usepackage[sc]{mathpazo}
\usepackage{geometry}
\usepackage{color}
\usepackage{array}
\usepackage{bbding}
\usepackage{multirow}
\usepackage[fleqn]{amsmath}
\usepackage{amssymb}
\usepackage{amsthm}
\usepackage{graphicx}
\usepackage{natbib}
\usepackage{lscape}
\usepackage[hidelinks]{hyperref}
\usepackage{comment}
\usepackage{bm}
\usepackage[title]{appendix}
\usepackage{longtable}
\usepackage{mathtools}
\usepackage{chngcntr}
\usepackage{authblk}
\usepackage{siunitx}
\usepackage{babel}
\usepackage{dsfont}
\usepackage[most]{tcolorbox}
\usepackage[short]{optidef}
\usepackage{algorithm}
\usepackage{algpseudocode}

\makeatletter
\allowdisplaybreaks
\usepackage[mathlines, pagewise]{lineno}
\usepackage{etoolbox} %% <- for \pretocmd, \apptocmd and \patchcmd

\newcommand*\linenomathpatchAMS[1]{%
	\expandafter\pretocmd\csname #1\endcsname {\linenomathAMS}{}{}%
	\expandafter\pretocmd\csname #1*\endcsname{\linenomathAMS}{}{}%
	\expandafter\apptocmd\csname end#1\endcsname {\endlinenomath}{}{}%
	\expandafter\apptocmd\csname end#1*\endcsname{\endlinenomath}{}{}%
}

\expandafter\ifx\linenomath\linenomathWithnumbers
\let\linenomathAMS\linenomathWithnumbers
\patchcmd\linenomathAMS{\advance\postdisplaypenalty\linenopenalty}{}{}{}
\else
\let\linenomathAMS\linenomathNonumbers
\fi

\linenomathpatchAMS{gather}
\linenomathpatchAMS{multline}
\linenomathpatchAMS{align}
\linenomathpatchAMS{alignat}
\linenomathpatchAMS{flalign}

\@ifundefined{showcaptionsetup}{}{%
 \PassOptionsToPackage{caption=false}{subfig}}
\usepackage{subfig}
\makeatother

\usepackage{babel}

\def\d{\mathop{\rm \!d}\!}

\usepackage{cleveref}
\usepackage{amsmath} % For \text command and better math formatting
\usepackage{amssymb} % For symbols like \ge
\usepackage{booktabs}

\usepackage{hyperref}

\begin{document}
	%\title{Learning to Align LLM with Human Travel Choices: Adaptive Persona Conditioning with Embedding Representations}
	%\title{Learning to Align LLM with Human Travel Choices via Personas and Embedding Representations}
    \title{Aligning LLM with human travel choices: a persona-based embedding learning approach}
	\author[1]{Tianming Liu}
    \author[1]{Manzi Li}
	\author[1]{Yafeng Yin\footnote{Corresponding author. E-mail address: \textcolor{blue}{yafeng@umich.edu} (Y. Yin).}}

	\affil[1]{\small\emph{Department of Civil and Environmental Engineering, University of Michigan, Ann Arbor, United States}\normalsize}
 %\affil[2]{\small\emph{Department of Electrical Engineering and Computer Science, University of Michigan, Ann Arbor, United States}\normalsize}
	\date{\today}
	\maketitle

\begin{abstract}
     %The advent of large language models (LLMs) presents new opportunities for travel demand modeling, as they can engage in the transportation planning process by acting as simulated entities of human travelers. While the approach has great potential, however, significant hurdles remain: current LLMs often exhibit behavioral misalignment with humans, and existing methods to address this are frequently inefficient or impractical given the constraints of typical travel demand data. 
     The advent of large language models (LLMs) presents new opportunities for travel demand modeling. However, behavioral misalignment between LLMs and humans presents obstacles for the usage of LLMs, and existing alignment methods are frequently inefficient or impractical given the constraints of typical travel demand data. This paper introduces a novel framework for aligning LLMs with human travel choice behavior, tailored to the current travel demand data sources. Our framework uses a persona inference and loading process to condition LLMs with suitable prompts to enhance alignment. The inference step establishes a set of base personas from empirical data, and a learned persona loading function driven by behavioral embeddings guides the loading process. We validate our framework on the Swissmetro mode choice dataset, and the results show that our proposed approach significantly outperformed baseline choice models and LLM-based simulation models in predicting both aggregate mode choice shares and individual choice outcomes. Furthermore, we showcase that our framework can generate insights on population behavior through interpretable parameters. Overall, our research offers a more adaptable, interpretable, and resource-efficient pathway to robust LLM-based travel behavior simulation, paving the way to integrate LLMs into travel demand modeling practice in the future.
     
     %combines behavior science, representation learning, and prompt engineering, and comprises two core stages: initially, we use expert LLMs to infer rich, interpretable textual personas from observational data to enrich the behavior context. Following this, a novel persona loading function, driven by representation learning, is learned via semi-supervised learning and leads the conditioning of a downstream LLM through dynamic, structured prompts, thereby targeting precise behavioral alignment. We validate our framework on the Swissmetro mode choice dataset, and the results show that our proposed approach significantly outperformed benchmark choice models and LLM-based simulation models in predicting both aggregate mode choice shares and individual choice outcomes. Furthermore, we showcase that our framework can generate additional insight on population behavior through interpretable parameters. Overall, our research offers a more adaptable, interpretable, and resource-efficient pathway to robust LLM-based travel behavior simulation, paving the way to integrate LLMs into travel demand modeling practice in the future.
\end{abstract}

\hfill\break%
	
%\begin{abstract}
%\noindent In transportation system demand modeling and simulation, agent-based models and microsimulations are current state-of-the-art approaches. However, existing agent-based models still have some limitations on behavioral realism and resource demand that limit their applicability. In this study, leveraging the emerging technology of large language models (LLMs) and LLM-based agents, we propose a general LLM-agent-based modeling framework for transportation systems. We argue that LLM agents not only possess the essential capabilities to function as agents but also offer promising solutions to overcome some limitations of existing agent-based models. Our conceptual framework design closely replicates the decision-making and interaction processes and traits of human travelers within transportation networks, and we demonstrate that the proposed systems can meet critical behavioral criteria for decision-making and learning behaviors using related studies and a demonstrative example of LLM agents' learning and adjustment in the bottleneck setting. Although further refinement of the LLM-agent-based modeling framework is necessary, we believe that this approach has the potential to improve transportation system modeling and simulation.
%\end{abstract}
%\hfill\break%
%\noindent\textit{Keywords} -Travel demand modeling, Large-language model, Agent-based simulation, Transportation planning\normalsize

\newpage

\section{Introduction}
%In travel demand analysis, modeling and predicting travel choice behavior, such as departure time choice, mode choice, and route choice, are critical for transportation planners in the planning process. For this task, discrete choice models (DCMs), grounded in microeconomic utility theory and offering high interpretability, have long served as benchmarks for analysis worldwide. However, in practice, DCMs require \textit{a-priori} specification of a fixed set of parameters and variables, limiting the direct inclusion of factors outside this predefined structure and tying the model's scope closely to the specific context for which it was calibrated. Furthermore, estimating each DCM necessitates a calibration process using collected structured cross-sectional data, which can be resource-intensive. Although machine learning models have been utilized to increase the expressiveness of DCMs, the limitations mentioned above persist.

In travel demand analysis, modeling and predicting travel choice behavior, such as departure time choice, mode choice, and route choice, are critical cornerstones for informed decision-making in transportation planning. Recently, the emergence of large language models (LLMs) presents new opportunities to expand and enhance the modeling of travel choice behavior. Pre-trained using billions of parameters on vast and diverse corpora, contemporary LLMs such as ChatGPT \citep{achiam2023gpt}, Gemini \citep{team2023gemini}, Claude \citep{anthropic2024claude}, Deepseek \citep{liu2024deepseek}, and Llama \citep{touvron2023llama} have been shown to possess unprecedentedly strong representation power in imitating human-generated patterns and generating human-like contents. Furthermore, they possess the ability to leverage unstructured qualitative inputs and natural language instructions, and have emerging capabilities in contextual understanding and complex task completion with limited \citep{brown2020language} to no \citep{kojima2022large} demonstration. These foundational capabilities motivate a central premise of our research: an LLM, when effectively aligned with human travel behavior through appropriate conditioning, can serve as a powerful base model to contribute to travel demand modeling practices. The LLMs' versatility in specification via natural language prompting can enable flexible model specification. Furthermore, their base training data and capacity to handle diverse input types may assist the model calibration process, and their representation power may enable models to generalize to new choice contexts \citep{liu2024toward}. Promising new applications with LLMs, such as exploratory scenario exploration, hypothesis generation, and policy screening, have the potential to greatly enhance travel choice modeling and transportation planning practices.

However, currently, utilizing LLMs for travel choice behavior simulation still faces significant challenges. While established data collection procedures center on utilizing choice contexts and socio-demographic variables as primary inputs of choice models, extensive research in LLM-based simulation in both travel choice contexts \citep{liu2024can} and other social science contexts (e.g., \cite{chen2023emergence,hagendorff2023human,goli2023can,park2024diminished}) indicates that these variables alone are often insufficient to steer LLMs towards accurately replicating human choice patterns. Moreover, existing alignment methods, such as supervised fine-tuning and conventional persona-based methods, all present significant limitations regarding computational demands and practical compatibility with typical travel demand datasets. Consequently, to unlock the potential of LLMs, there is a clear need for new mechanisms that can effectively align LLM-generated travel choices with those of human travelers and be compatible with the data and resource landscapes of travel demand modeling.

In this paper, we propose a novel learning approach for LLM alignment tailored for the typical travel demand data landscape. To achieve alignment, our method conditions LLMs with personas, whose selection mechanism is determined through representation learning of embeddings. Our key idea is twofold: First, we infer personas of a subset of travelers and use them as core elements to guide LLM's behavior through prompting. Second, informed by the correlation between socio-demographics and travel behavior, we employ representation learning to estimate a persona loading function that controls the prompt generation process through learning the behavioral representations of socio-demographic groups. The resulting framework provides a powerful tool to transform survey data into LLM prompts to accurately simulate traveler behavior, and furthermore, can be used to extract additional behavioral insights through interpretations of the learned model parameters. We empirically tested our framework using the Swissmetro dataset, benchmarking its performance in predicting both aggregate choice distributions and individual travel outcomes against benchmark discrete choice models and established lightweight LLM alignment techniques. The evaluation demonstrates that our proposed method achieves superior performance at both the aggregated alternative shares and individual behavior prediction levels, highlighting its significant efficacy and potential for advancing behavioral modeling.

To our knowledge, our paper presents the first learnable prompt-based approach for LLM travel behavior alignment. Methodologically, our framework integrates behavioral foundations, representation learning and LLM in a novel fashion, allowing each element to play to its strengths. Therefore, our approach achieves superior performance than existing alignment methods in travel demand modeling tasks. Practically, our approach enables modelers and practitioners to directly leverage ``off-the-shelf" low-cost LLMs for behavioral simulations without requiring extensive hardware and data for supervised fine-tuning. This makes our approach a significant step towards more accessible and robust LLM tools for travel demand analysis and transportation planning.

The remainder of this paper is outlined as follows: first, \Cref{sec:LR} reviews relevant literature in LLM alignment and representation learning. Then, \Cref{sec:preliminaries} sets up the stage by mathematically introducing the alignment problem and contexts, and \Cref{sec:method} details our methodological framework and its theoretical properties. The setting and results of our empirical experiment are elaborated in \Cref{sec:experiment} and \Cref{sec:results}, respectively. Finally, \Cref{sec:conclusion} concludes the paper.

\section{Related Work} \label{sec:LR}
Our work is related to two streams of research in LLM and one research trend in applying representation learning in travel demand modeling. In this section, we review related literature and position our work in each stream.

\subsection{LLM-based behavior simulation}

In the last few years, the advancement of LLMs has propelled increasing research into their application for simulating human behavior within complex social systems \citep{grossmann2023ai,park2023generative,gao2024large,anthis2025llm}. Owing to their training on gigantic sets of human-generated data and their extensive parameter spaces, LLMs exhibit unprecedented capacities to generate human-like responses in the form of natural language across diverse contexts. This inherent ability positions them as powerful tools for behavioral simulation, leading researchers to explore their potential in mitigating longstanding challenges in survey research, such as rising costs and increasing non-response bias \citep{jansen2023employing,ziems2024can,anthis2025llm} and act as ``guinea pigs" in policy simulations to produce assessments with greatly flexibility and generalization \citep{grossmann2023ai,argyle2023out,liu2024toward}. Indeed, empirical studies indicate that LLMs demonstrates some traits similar to humans when prompted to simulate human behavior \citep{argyle2023out,aher2023using,horton2023large,korinek2023language,demszky2023using,park2023generative,ziems2024can}.

However, despite the significant potential of LLMs in behavior simulation, correctly conditioning them to produce accurate simulations results is a challenging task \citep{tseng2024two}. Existing methods in travel demand modeling use contexts and socio-demographic characteristics as the main predictors of human behavior. However, providing this information to an LLM through role-playing prompts \citep{argyle2023out} has been proven to be generally insufficient. Extensive research in transportation \citep{liu2024can,chen2024delayptc} and broader social sciences \citep{hagendorff2023human,goli2023can,chen2023emergence,tjuatja2024llms,park2024diminished,hu2024quantifying,wang2024large,fan2024can,agnew2024illusion,beck2024sensitivity,li2025llm,chen2025perceptions} reveals that such direct prompting fails to reliably condition LLMs to produce human-like behavior across a wide range of contexts. Therefore, to fully unlock the potential of LLMs for advancing travel demand modeling research, new methodologies are necessary to enhance LLM conditioning and achieve closer alignment with actual human travel behavior. In response to this need, our work details a new framework to robustly condition LLMs for travel behavior simulation.

\subsection{LLM alignment} \label{subsec:LLM_alignment}
Aligning LLMs with human behavior patterns requires mechanisms to steer LLMs towards desirable behaviors. Overall, the mechanisms in established literature can be categorized into three prominent approaches: few-shot learning, supervised fine-tuning, and persona-based conditioning. Their mechanisms, strengthen and drawbacks are outlined below:

The first common strategy, few-shot learning, leverages an LLM's ability to generalize from a small number of examples provided directly within the prompt \citep{brown2020language}. In this approach, several examples of contextualized human behavior are supplied to the LLM, which is instructed to learn patterns from these examples and apply them to new contexts to simulate human behavior. While this approach is relatively straightforward to implement and has achieved notable success in various empirical studies \citep{kim2024learning,liu2024can}, it also faces significant limitations. Firstly, its heavy reliance on the LLM's information processing and inference capabilities can cap its performance \citep{song2023comprehensive,chamieh2024llms}. Secondly, it often exhibits diminishing returns to scale, and at times, increasing the number of examples can even be detrimental to performance \citep{liu2024can}. Furthermore, a practical challenge of the few-shot learning approach is that few-shot examples can be very long in token size, leading to substantially longer input prompts, which in turn can reduce inference speed and increase operational costs.

The second approach, supervised fine-tuning, utilizes supervised learning to further train a pre-trained LLM on a dedicated dataset of observed human behaviors and their corresponding contexts. Through this process, the LLM's parameters are adjusted (typically via backpropagation) to optimize its ability to replicate the target behaviors, often by maximizing the likelihood of generated behavior compared with the human ground truth. SFT has demonstrated good efficacy in increasing the behavioral alignment of LLMs in established studies \citep{binz2024centaur,liu2024nextlocllm,gong2025mobility,lu2025beyond}. Nevertheless, SFT is computationally intensive, necessitating significant hardware resources and access to extensive, high-quality labeled datasets for successful implementation. Without these prerequisites, SFT can be inefficient \citep{chuang2024beyond} and may even induce adverse effects such as catastrophic forgetting of prior knowledge \citep{zhai2023investigating}.

The third strategy centers on persona-based conditioning of LLMs. In this approach, LLMs are provided with personas, which are concise representations of individuals' preferences, behavioral traits, or other characteristics not explicitly captured by standard socio-demographic data. By incorporating these personas into role-play prompts, LLMs can gain a richer understanding of the behavior of individuals, leading to more accurate behavioral simulations. The persona-based approach has shown success in LLM alignment tasks within both transportation contexts \citep{wang2024large,liu2024can,li2024more} and other domains \citep{chuang2024beyond,park2024generative,chen2024persona,sun2024persona,chen2025perceptions}. However, in established approaches, the persona data are either established by specialized survey data \citep{chuang2024beyond,park2024generative,chen2025perceptions} or inferred from detailed cross-sectional data from the population level \citep{wang2024large,li2024more,sun2024persona}. Such comprehensive data sources are not usually available in travel demand research, thereby limiting the applicability of existing models in our context.

Motivated by the limitations of existing methods, in this paper, we propose an alignment method tailored to the data landscape of travel demand modeling. Inspired by insights from behavior theory and discrete choice modeling, our proposed approach relaxes the reliance on large-scale cross-sectional data of existing persona-based alignment methods and allows accurate alignment with more typically available sampling data. Balancing representation power with computational demand, our approach is lightweight in nature and can be run with standard computing tools or APIs. Our offers improved performance over few-shot learning while being significantly less resource-demanding than SFT, thereby enhancing accessibility and providing a robust tool for a wide range of practical applications in travel demand research.

\subsection{Representation learning in travel demand modeling}

Over the recent years, a stream of research on travel demand modeling has explored the application and integration of representation learning and discrete choice modeling. Combining the behavioral foundation of microeconomics and the expressivity of representation learning, this research endeavor seeks to enhance the predictive power of travel choice models through learning representations of travelers and travel contexts. Representation learning techniques such as fully-connected neural networks \citep{wang2020deep1,wang2020deep2,wang2021deep,han2022neural,feng2024agentmove,haj2025incorporating}, convolutional neural networks \citep{sifringer2020enhancing,lahoz2023attitudes}, residual neural networks \citep{wong2021reslogit,wang2021theory}, embedding representations \citep{pereira2019rethinking,arkoudi2023combining,wang2024deep,yang2024applying}, and graph neural networks \citep{ma2025incorporating} have been applied in the structure of predictive models of discrete choices. Among these approaches, embedding representations are most relevant to our work. Such embeddings function as learnable projections, mapping high-dimensional input spaces to more compact, latent representations. Prior studies have leveraged these for various purposes in modeling travel behavior: \cite{pereira2019rethinking} and \cite{arkoudi2023combining} utilized learnable embedding matrices to generate compact representations of discrete variables in a choice problem for integration into the utility function; \cite{wang2024deep} used a variational autoencoder to synthesize and extract information from figures to enrich the context; and \cite{yang2024applying} utilized natural language embeddings that condenses the meaning of textual representations to facilitate mode choice prediction. 

Our work diverges from these precedents primarily in its overall learning approach and the functional role of representation learning, particularly through a novel integration of representation learning and LLMs. While existing studies typically employ supervised learning models with each trained on specific choice settings and fixed sets of alternatives, our approach integrates LLM with representation learning in simulating the choices of human travelers to create a model that is flexible in context specification and possess potential in generalization. Furthermore, we introduce a distinct application of embedding representations: rather than primarily serving as direct inputs to utility calculations, we use them as a tool to explicitly encode and discover latent behavioral similarities between population groups, thereby enriching the application of representation learning in the travel choice modeling domain.

%\subsubsection{Problem definition}

%For the structure of $\Tilde{X}(d,X)$, we make the following two assumptions:
%\begin{assumption}
%$\Tilde{X}(d,X)=(d,X)$ is insufficient to achieve alignment.
%\label{assumption:zero_shot_insufficiency}
%\end{assumption}

%Assumption \ref{assumption:zero_shot_insufficiency} can be %justified by various existing evidence. 

\section{Preliminaries} \label{sec:preliminaries}
In this section, we present the definition and data context for the LLM alignment task. Our discussion centers on the LLM alignment problem in the context of discrete choices in travel demand modeling, and the overall objective is to condition the LLM so that its travel choices are consistent with human travelers' travel choices. The conditioning is achieved by prompting while keeping the LLM's weights frozen. The alignment problem necessitates observations of human travelers' choice behavior. Following current mainstream approaches of data collection in travel demand modeling, we assume that only the social-demographic information of the human traveler, the choice context, and the choice of the human traveler are available in such observation to maintain the generality of our approach. Below, we formally define the alignment problem and the datasets available for the alignment task.

\begin{comment}

\begin{table}[h]
  \centering
  \begin{tabular}{lll}
    \toprule
    Symbol & Description & Dimension / Domain \\
    \midrule
    $k$ & Individual index in high‑fidelity panel & $1 \le k \le K_h$ \\
    $i$ & Entry index in low‑fidelity survey & $1 \le i \le N_\ell$ \\
    $\mathbf \boldsymbol{d_k}, \mathbf d_i$ & Socio‑demographic vector & $\in \mathbb R^{D_d}$ \\
    $\mathbf \boldsymbol{X_k^j}, \mathbf \boldsymbol{X_i}$ & Choice‑context features & $\in \mathbb R^{D_x}$ \\
    $Y_k^j, Y_i$ & Observed choice (alternative index) & $\in \{1,\ldots,M\}$ \\
    $J_k$ & \# of repeated observations for individual $k$ & positive integer \\
    $\widetilde{\mathbf X}(d,X)$ & Persona prompt generated for the LLM & $\in \mathbb R^{D_t}$ \\
    $\mathrm{LLM}(\cdot)$ & Language‑model predictor & $\mathbb R^{D_t}\to\{1,\dots,M\}$ \\
    \bottomrule
  \end{tabular}
  \caption{Key notation used throughout.}
  \label{tab:notation}
\end{table}
\end{comment}

\subsection{The LLM alignment problem} \label{subsec:prem_alignment_definition}

Consider the population and travel choice behaviors of interest, which we characterized by an underlying distribution $\mathcal{P}(\boldsymbol{d},\boldsymbol{X},Y)$ with pdf $f_{\mathcal{P}}$:
\begin{itemize}
    \item $\boldsymbol{d}\in \mathbb{R}^{D_d}$ is the social-demographic information of the respondent, characterized by a vector of dimension ${D_d}$.
    \item $\boldsymbol{X} \in \mathbb{R}^{D_X}$ is the context of the choice problems faced by the respondent, characterized by a vector of dimension $D_X$.
    \item $Y$ is the chosen alternative by the respondent.
\end{itemize}

In our alignment framework, we allow the inclusion of multiple behaviors within the choice context. Therefore, $\boldsymbol{X}$ and $Y$ can be a mixture of multiple behaviors, such as a mixture of vehicle ownership and mode choices. Correspondingly, we do not assume that the choice $Y$ follows a uniform choice set and without loss of generality, we set $D_x$ to be large enough to encompass information from all contexts. 

For alignment, we condition an LLM model, denoted by $LLM$, to simulate the population's behaviors of interest $\mathcal{P}(\boldsymbol{d},\boldsymbol{X},Y)$. The LLM acts as a simulated entity of a human traveler, reacting to an input prompt and outputting the simulated choice:
\begin{equation}
    \hat{Y}=LLM(G_{\Phi}(\boldsymbol{d},\boldsymbol{X}))
    \label{eq:basic_llm_role_play}
\end{equation}
\noindent in which $G_{\Phi}(\boldsymbol{d},\boldsymbol{X})$ is a prompt generation function conditioned on socio-demographics $\boldsymbol{d}$ and choice context $\boldsymbol{X}$, $\Phi$ is the parameter of the prompt generation function, and $\hat{Y}$ is the alternative chosen by the LLM. The output of $G_{\Phi}$ is a textual prompt that is a series of tokens and does not have a fixed dimension. 

Given this LLM usage, the alignment problem can be mathematically expressed as:

\begin{equation}
\max_{\Phi} \sum_{Y \in \mathcal{Y}} \iint_{\mathcal{X}, \mathcal{D}} \left( \log \mathbb{P}[LLM(G_{\Phi}(\boldsymbol{d},\boldsymbol{X}))=Y] \right) f_{\mathcal{P}}(\boldsymbol{d},\boldsymbol{X},Y) \d\boldsymbol{d} \d\boldsymbol{X}
\label{eq:population_level_alignment_problem}
\end{equation}

%\begin{equation}
%    \max_{\Phi} %\mathbb{E}_{(\boldsymbol{d},\boldsymbol{X},Y)\sim\mathcal{P}}[log\:\mathbb{P}(LLM(G_{\Phi}(\boldsymbol{d},\boldsymbol{X}))=Y)]
   % \label{eq:population_level_alignment_problem}
%\end{equation}
\noindent which maximizes the expected log-likelihood that the LLM would generate the human travelers' actual choices over the distribution $\mathcal{P}$.

The alignment problem described in \Cref{eq:population_level_alignment_problem} can be seen as finding the optimal prompt-generating function $G_\Phi$ to generalize in the population, which presents notable challenges. First, the training of $G_\Phi$ lacks explicit supervised labels because the prompts are not known in the training data, therefore complicating the estimation process. Secondly, the textual nature of $G_\Phi$'s output significantly complicates learning, as the text space has high and varied dimensions, making specification and estimation of $G$ more difficult. Addressing these challenges while maintaining the computational framework lightweight and accessible is 
a critical objective for our methodology.

%In travel‑behavior modeling, detailed panel surveys (high‑fidelity) capture repeated choices for the same individuals, whereas large‑scale household surveys (low‑fidelity) record only one choice per respondent. Our aim is to align an LLM’s predictions with human decisions: given socio‑demographics $\boldsymbol{d}$ and context $\boldsymbol{X}$, the LLM should output the same travel choice $Y$ as a human. To achieve this, we learn a \emph{conditional‑input generator}
%\[
%  \widetilde X_\phi(d,X)\;:\;\mathbb R^{D_d}\times\mathbb R^{D_x}\;\longrightarrow\;\mathbb R^{D_t},
%\]
%which produces a tailored persona prompt. Training leverages both datasets so that
%\[
%  \mathrm{LLM}\bigl(\widetilde X_\phi(d,X)\bigr)
%  \;\approx\;Y,
%\]
%thereby ensuring behavioral alignment between model and traveler.

\subsection{Data sources} \label{subsec:prem_dataset}

In the alignment task, we assume access to two distinct datasets $\mathcal{D}_h$ and $\mathcal{D}_\ell$, both of which represent data types commonly seen in the current travel demand datasets:

\begin{enumerate}
  \item \textbf{Detailed observational data}
    \[
      \mathcal{D}_h
      = \Bigl\{\,\bigl( \boldsymbol{d_k},\{( \boldsymbol{X_k^j},Y_k^j)\}_{j=1}^{J_k}\bigr)
      \;\big|\;k=1,\ldots,K_h\Bigr\}.
    \]
    $\mathcal{D}_h$ contains detailed choice behavior observations from $K_h$ persons from the population. For each individual $k$, their socio-demographic information $\boldsymbol{d_k}$ and their choices $Y_k^1,Y_k^2,...,Y_k^{J_k}$ under $J_k$ different contexts $X_k^1,X_k^2,...,X_k^{J_k}$ are recorded in the dataset. In practice, $\mathcal{D}_h$ corresponds to targeted data such as research-oriented stated-preference surveys or revealed preferences from a recruited group of participants.

  \item \textbf{General observational data}
    \[
      \mathcal{D}_\ell
      = \Bigl\{\,( \boldsymbol{d_i}, \boldsymbol{X_i},Y_i)\;\big|\;i=1,\ldots,N_\ell\Bigr\}.
    \]

    $\mathcal{D}_\ell$ contains observations of $N_\ell$ instances of choice behavior from the overall population. Each observation in this dataset is denoted as $i$, and contains the corresponding socio-demographics $\boldsymbol{d_i}$, choice context $\boldsymbol{X_i}$ and chosen alternative $Y_i$. However, unlike $D_h$, for each individual here, we do not assume that we can observe their choice behavior under different contexts because such general observational data are usually sparse or have been anonymized. In practice, this dataset corresponds to general-purpose data, such as general travel surveys or large-scale revealed preference data, such as GPS data, both are usually anonymized. A part of a larger targeted data can also serve as $\mathcal{D}_\ell$.
\end{enumerate}

\begin{comment}
\begin{figure}
    \centering
    \includegraphics[width=0.8\linewidth]{data and alignment.png}
    \caption{data}
    \label{fig:enter-label}
\end{figure}
\end{comment}

\section{Methodology} \label{sec:method}

In accordance of the premise in \Cref{sec:preliminaries}, we propose an alignment method combining behavior theory, prompt engineering, LLM inference, and representation learning to regulate and estimate $G_\Phi$ for the alignment problem \ref{eq:population_level_alignment_problem}. Overall, our approach encompasses two main aspects:
\begin{itemize}
    \item The first aspect involves structuring the output of the prompt generator. Using insights from behavior theory, we use structured prompting to split the prompt into a context-insensitive backbone and context-varied inputs, and identify a persona that contains the traveler's preferences as the additional element of context-sensitive inputs alongside socio-demographics and choice context. Furthermore, we obtain labels of the persona by inferring from the detailed observational data $\mathcal{D}_h$ in order to create a behaviorally grounded basis for prompt generation.

    \item The second aspect of our approach focuses on specifying and estimating the prompt generator. Our main intuition is twofold: first, the prompt generator $G_\Phi$ should be probabilistic due to the heterogeneity in behavior among travelers in the same population group; and second, this probabilistic generation process can be steered by latent behavioral similarities between population groups due to the correlation between preferences and socio-demographic characteristics. Therefore, we propose a representation learning approach to optimize the prompt generator via learning the latent representation of socio-demographics in an embedding space that measures their behavioral similarity. Furthermore, we utilize a Monte-Carlo stochastic expectation-maximization (EM) algorithm for the learning process to reduce the cost of LLM inference calls, making the proposed framework more accessible.
\end{itemize}

\begin{figure}[h!]
    \centering
    \includegraphics[width=1\linewidth]{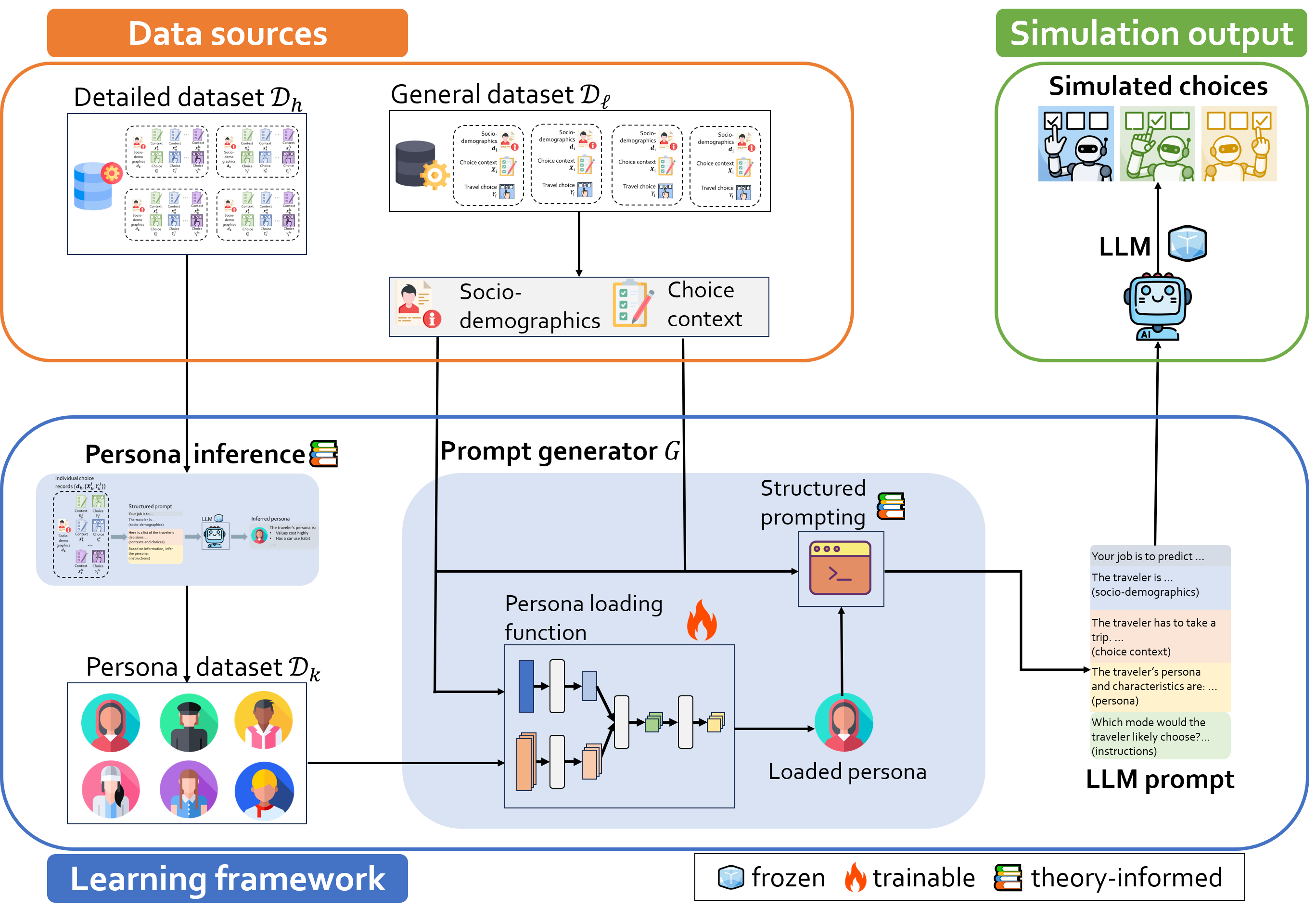}
    \caption{Overview of our alignment framework}
    \label{fig:overview_framework}
\end{figure}

An overview of the proposed learning process and its two main components is shown in \Cref{fig:overview_framework}. We elaborate on the design and implications of each component in the following subsections.

\subsection{Step 1: prompt regularization and labeling with persona information} \label{subsec:method_prompt_structure}

A key task for the prompt generator $G$ is to generate prompts that allows the LLM to accurately output the choice of the traveler. Therefore, the prompt must contain sufficient information about the overall travel situation, the choice context, the traveler, and the LLM's task to allow the LLM to understand the background information and output related choices. To achieve this, we utilize a structured prompting approach to regulate the prompt generated by $G$. Each generation prompt is composed of two primary components: firstly, a static, context-insensitive backbone that establishes a consistent task definition and provides general operational instructions to the LLM; and a set of context-varied inputs that reflect the specific context of each choice setting. This structured prompting approach provides an explicit mechanism for integrating key information directly into the LLM's input stream, thereby enhancing its ability to accurately interpret context and generate choices.

Regarding the necessary information in the context-varied inputs of the prompt, we define them by drawing insights from microeconomics and discrete choice models (DCMs). In discrete choice modeling, three pieces of information are generally contained in the formulation of the utility function and choice model: (1) the choice context, (2) the socio-demographic characteristics of the respondent, and (3) parameters representing the individual's economic preferences and behavioral traits, and all three components are necessary in predicting an individual's choice behavior. Accordingly, we design the context-varied portion of our prompt to feature these three components. Consequently, the LLM input prompt adheres to the format of:
\begin{equation}
    g(\boldsymbol{d},\boldsymbol{X},\boldsymbol{Z})
    \label{eq:input_prompt_format}
\end{equation}
\noindent in which $g$ denotes the prompt structure, $\boldsymbol{d} \in \mathbb{R}^{D_d}$ is the vector containing socio-demographic information, $\boldsymbol{X}\in \mathbb{R}^{D_X}$ is the vector about the choice context, and $\boldsymbol{Z}$ is defined as a persona that captures a traveler's economic preferences and behavioral traits. While both $\boldsymbol{d}$ and $\boldsymbol{X}$ are vectors, the format of $\boldsymbol{Z}$ is more flexible: it can be a vector akin to parameters in a discrete choice model, as well as a body of text description, the latter more commonly seen in LLM simulation literature \citep{wang2024large,li2024more,park2024generative,sun2024persona}. Thus, we do not impose assumptions of the format of $\boldsymbol{Z}$.

%The structured prompt format in \Cref{eq:input_prompt_format} does not only align with microeconomic theory and discrete choice modeling practices, but is further supported by their successful application in recent empirical studies focused on enhancing personalization and modeling complex agent behaviors \citep{argyle2023out,tseng2024two}.

\begin{figure}[h!]
    \centering
    \includegraphics[width=0.7\linewidth]{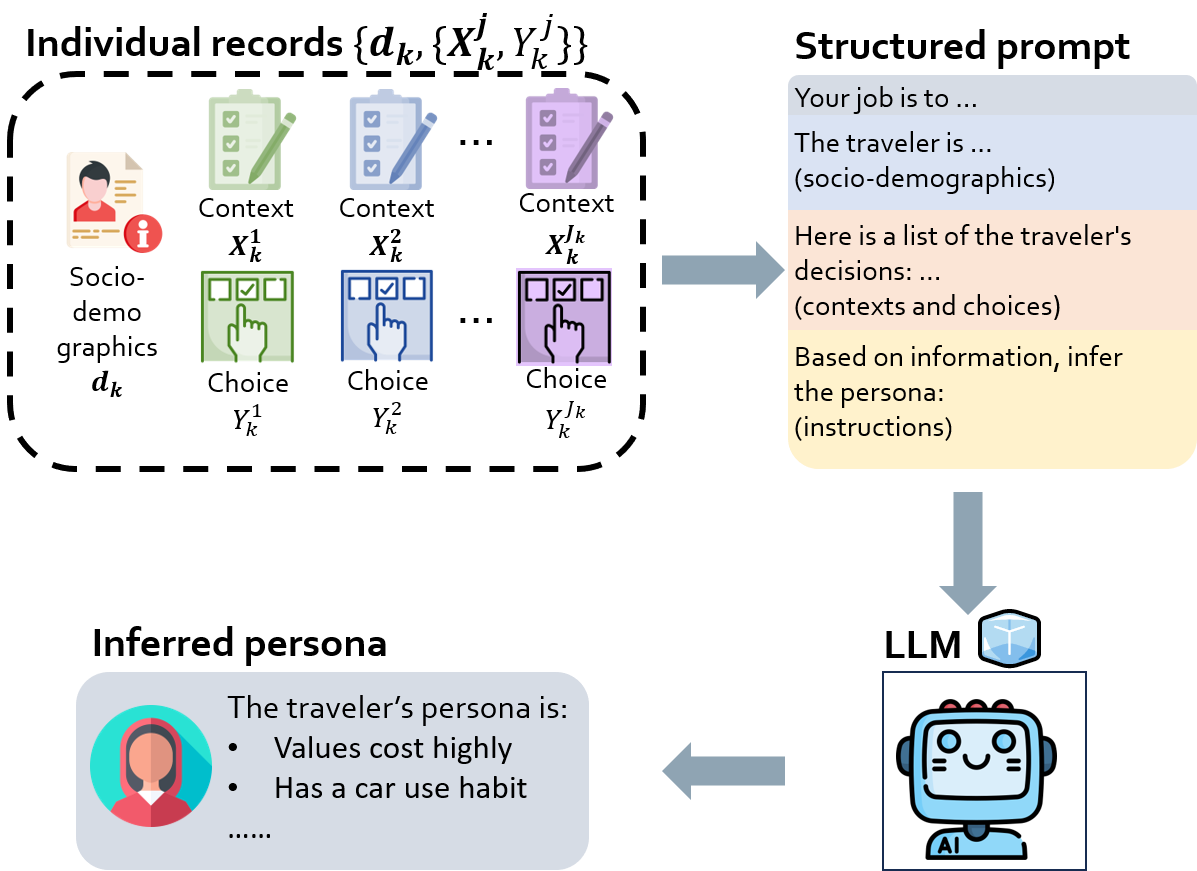}
    \caption{Illustration of the persona inference process}
    \label{fig:persona_inference_step}
\end{figure}

While the structured prompting approach greatly reduces the uncertainty and complexity of the input prompts, the challenges regarding the lack of labeling and fixed dimensions remains due to the anonymity and flexibility of $Z$. We address this challenge by leveraging the detailed observational data to establish a basis of personas that can be further used in the subsequent learning process. Specifically, for each individual $k$ in the detailed observation dataset $\mathcal{D}_h$, we employ an expert LLM (denoted by $LLM_e$) which is provided with the individual's socio-demograhic vector $\boldsymbol{d_k}$ along with the full sequence of $J_K$ observed context–choice pairs $\{(\boldsymbol{X_k^j},\, Y_k^j)\}_{j=1}^{J_k}$ for the individual to infer this person's persona:
\begin{equation}
     \boldsymbol{Z_k}=LLM_e\Bigl( \boldsymbol{d_k},\;\{(\boldsymbol{X_k^j},\,Y_k^j)\}_{j=1}^{J_k}\Bigr)
     \label{eq:persona_inference}
\end{equation}
\noindent the expert LLM's task is to infer the traveler's economic preferences and behavioral traits from their choice behavior and synthesize the inferred information into a personalized textual description $\boldsymbol{Z_k}$ (illustrated in \Cref{fig:persona_inference_step}).

Applying $LLM_e$ on every person $k\in \mathcal{D}_h$ in the detailed observation data, we obtain a collection of generated personas $\mathcal{D}_k=\{\boldsymbol{Z_k}\}_{k\in \mathcal{D}_h}$, which providing a concrete basis of persona that can guide the subsequent learning process. Beyond enhancing the feasibility of learning, our strategy of employing inferred personas also shares common ground with advanced discrete choice modeling practices. The inference of personas from detailed observations bears strong similarity with inferring the distributions of economic preference parameters, which is consistent with some discrete choice models such as the mixed logit model \citep{hensher2003mixed}. This dual justification from both the learning and behavior theory perspective positions this inference step as a crucial and robust foundation for the overall alignment process.

\subsection{Step 2: loading-based prompt generation} \label{subsec:method_matching}
After defining the format of the output of the prompt generator $G$ in \Cref{subsec:method_prompt_structure}, we specify the structure and the estimation process of the prompt generator in this section to complete the alignment process. For the alignment process, as the information in $\mathcal{D}_h$ has already been utilized by the persona inference step, we use the other dataset $\mathcal{D}_{\ell}$ to train the prompt generator to minimize the dissimilarity between LLM choices and human choices:
\begin{equation}
    \max_{\Phi} \sum_{(\boldsymbol{d_i},\boldsymbol{X_i},Y_i)\sim\mathcal{D}_{\ell}}[log\:\mathbb{P}(LLM(G_{\Phi}(\boldsymbol{d_i},\boldsymbol{X_i}))=Y_i)]
    \label{eq:population_level_alignment_problem_dataset}
\end{equation}
\noindent in which the prompt generator's output adheres to the format in \Cref{eq:input_prompt_format}.

By \Cref{eq:basic_llm_role_play} and \Cref{eq:input_prompt_format}, the prompt generator's output includes the persona $\boldsymbol{Z_i}$, which is correlated with the socio-demographics $\boldsymbol{d_i}$ but also heterogeneous within each population subgroup. To reflect this, we characterize the generator function by a discrete probability distribution $P(\boldsymbol{Z_k}|\boldsymbol{d_i})$ over the learned personas $\mathcal{D}_k$:
\begin{equation}
    \mathbb{P}(G_{\Phi}(\boldsymbol{d_i},\boldsymbol{X_i})=g(\boldsymbol{d_i},\boldsymbol{X_i},\boldsymbol{Z_k}))=P(\boldsymbol{Z_k}|\boldsymbol{d_i}) \quad \forall \boldsymbol{Z_k}\in \mathcal{D}_k
    \label{eq:stochastic_generator_function}
\end{equation}

Thus, for each observation of $\boldsymbol{d_i}$ and $\boldsymbol{X_i}$ in $\mathcal{D}_{\ell}$, the prompt generated by $G$ is regulated to instantiate the defined structure $g(\boldsymbol{d_i},\boldsymbol{X_i},\boldsymbol{Z_k})$ by stochastically matching a persona $\boldsymbol{Z_k}$ from the persona dataset $\mathcal{D}_k$ according to a persona loading function $P(\boldsymbol{Z_k}|\boldsymbol{d_i})$ that captures the relationship between personas and socio-demographic factors. By defining the prompt over the learned personas, we mitigate the challenges associated with the persona's inherent variability and its potentially high dimensionality, making the alignment process substantially more tractable and computationally efficient.

\Cref{eq:stochastic_generator_function} effectively transformed the alignment problem into estimating the optimal $P(\boldsymbol{Z_k}|\boldsymbol{d_i})$ to maximize the behavior similarity between the LLM and human travelers. In the following discussion, we first establish the structure of the persona loading function and then discuss the solution process of the alignment problem.

\subsubsection{Formulating the persona loading function}

\begin{figure}[h!]
    \centering
    \includegraphics[width=1\linewidth]{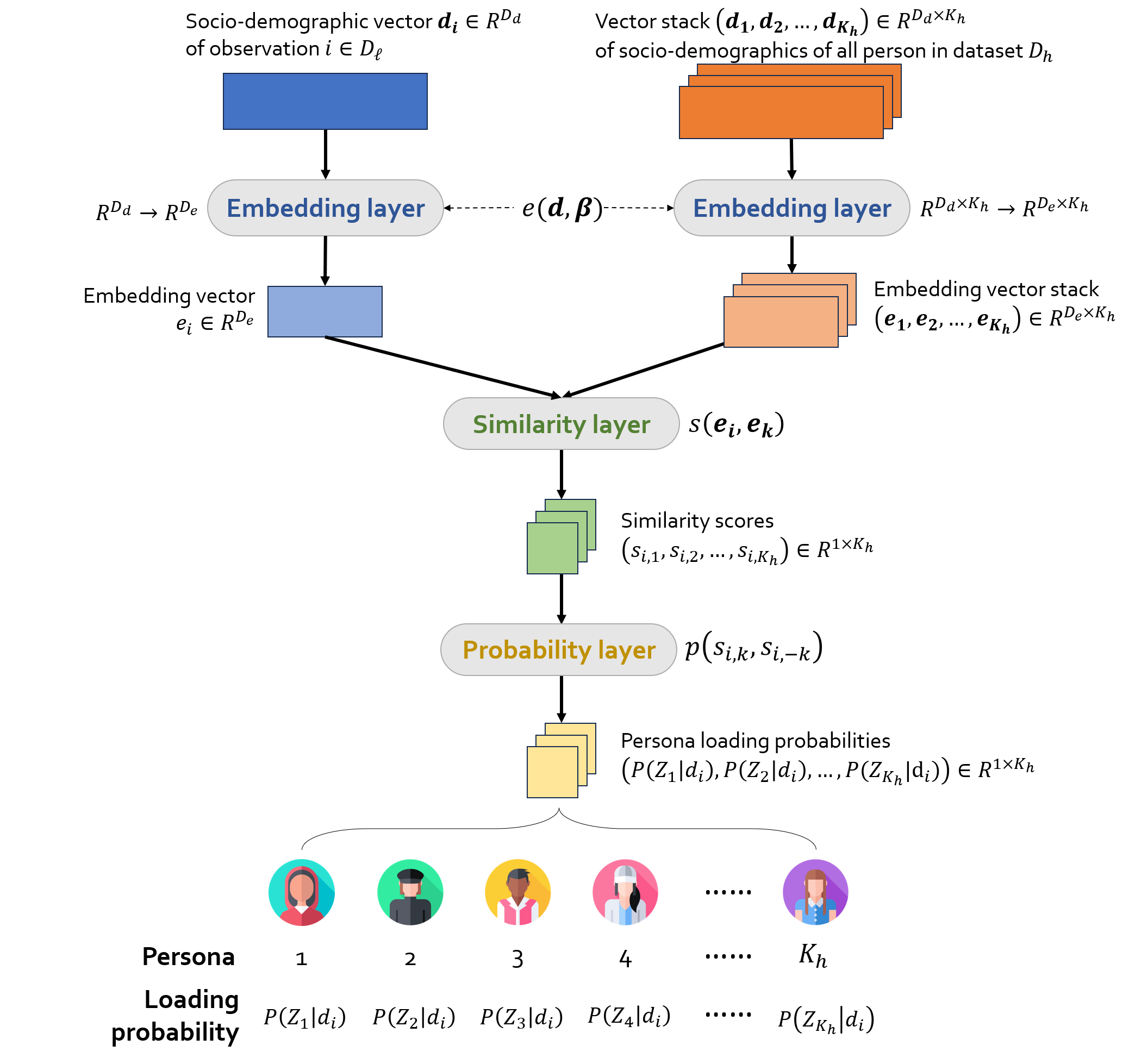}
    \caption{Illustration of the persona loading function}
    \label{fig:persona_loading_function}
\end{figure}

To define the persona loading function $P(\boldsymbol{Z_k}|\boldsymbol{d})$, our key idea is to exploit the behavior similarities between population groups and match personas more frequently from socio-demographic groups that are more behaviorally aligned with the target individual's profile. An overview of the persona loading function is shown in \Cref{fig:persona_loading_function}.

The behavioral similarity between groups is latent in nature and not directly observable from raw socio-demographic features. Therefore, we utilize an embedding function $e$ to establish the representation of socio-demographic groups in the latent space:
\begin{equation}
   \boldsymbol{e_i}=e(\boldsymbol{d_i};\boldsymbol{\beta})
    \label{eq:embedding_function}
\end{equation}
\noindent in which $\boldsymbol{\beta}$ is a set of learnable parameters that is invariant to $\boldsymbol{d_i}$, and $\boldsymbol{e_i}\in \mathbb{R}^{D_e}$, a $D_e$-dimension vector of continuous variables, is the embedding of $\boldsymbol{d_i}$. $e$ is also applied similarity to $\boldsymbol{d_k}$ to generate $\boldsymbol{e_k}$.

The embedding function $e:\mathbb{R}^{D_d}\rightarrow\mathbb{R}^{D_e}$ projects a socio-demographic vector $\boldsymbol{d_i}$ into a latent space wherein learned proximities between population subgroups are intended to capture their underlying behavioral affinities. Therefore, the similarity between the socio-demographics information $\boldsymbol{d_i}$ in observation $i\in \mathcal{D}_{\ell}$ and the socio-demographics information $\boldsymbol{d_k}$ of individual $k\in \mathcal{D}_h$ can be quantified using their embedding representation $\boldsymbol{e_i}$ and $\boldsymbol{e_k}$. We formalize this similarity measure using a scalar-valued scoring function $s$:
\begin{equation}
    s_{i,k}= s(\boldsymbol{e_i},\boldsymbol{e_k})
    \label{eq:similarity_score}
\end{equation}
\noindent in which $s_{i,k}$ is a scalar that measure the similarity between $\boldsymbol{d_i}$ and $\boldsymbol{d_k}$. This scoring function $s$ is designed such that a higher value of $s_{i,k}$ signifies a greater degree of similarity between the latent representations $\boldsymbol{e_i}$ and $\boldsymbol{e_k}$, and by extension, reflects a stronger alignment in their underlying behavioral characteristics. Furthermore, a well-behaved similarity function adheres to several key properties outlined below:
\begin{subequations}\label{eq:similarity_properties}
\begin{align}
    &\begin{aligned} % This inner 'aligned' environment preserves your two-line structure for each condition
        &\text{Condition 1: Boundedness}\\
        &|s(\boldsymbol{e_i},\boldsymbol{e_k})| <\infty
     \end{aligned} \label{eq:sim_non_negativity_bound} \\
    &\begin{aligned}
        &\text{Condition 2: Symmetry}\\
        &s(\boldsymbol{e_i},\boldsymbol{e_k}) = s(\boldsymbol{e_k},\boldsymbol{e_i})
     \end{aligned} \label{eq:sim_symmetry} \\
    &\begin{aligned}
        &\text{Condition 3: Self-similarity}\\
        &s(\boldsymbol{e_i},\boldsymbol{e_i}) \ge s(\boldsymbol{e_i},\boldsymbol{e_k})
     \end{aligned} \label{eq:sim_self_similarity}
\end{align}
\end{subequations}
\noindent Among the three critical conditions defined above, boundedness (Condition~\ref{eq:sim_non_negativity_bound}) ensures that similarity scores are finite, which is crucial for stable computations. Symmetry (Condition~\ref{eq:sim_symmetry}) reflects the notion that similarity is a mutual, order-independent relationship. Finally, self-similarity (Condition~\ref{eq:sim_self_similarity}) establishes that any representation is at least as similar to itself as it is to any other representation, adhering to the projection space interpretation and providing an essential reference point for comparisons.

Finally, to operationalize our core idea on the matching function, we now define the persona loading function using the established similarity measure $s$ as follows:
\begin{equation}
    P(\boldsymbol{Z_k}|\boldsymbol{d_i})=p(s_{i,k},s_{i,-k})
    \label{eq:loading_prob_dist}
\end{equation}
\noindent in which $s_{i,k}$ is the similarity measure between $\boldsymbol{d_i}$ and $\boldsymbol{d_k}$, $s_{i,-k}=(s_{i,1},s_{i,2},...,s_{i,k-1},s_{i,k+1},...,s_{i,K_h})$ is a vector containing the similarity measure between $\boldsymbol{d_i}$ and $\boldsymbol{d}$ of every person in $\mathcal{D}_h$ apart from person $k$, and $p:\mathbb{R}^{K_h}\rightarrow\mathbb{R}$ is a function mapping the similarities scores between the socio-demographics of $\boldsymbol{d_i}$ and the persons in $\mathcal{D}_h$ to a scalar value. The function $p$ connects similarity scores with the persona loading probability, and assigns higher probabilities for personas that are in population groups with higher underlying similarity in behavior. Therefore, it should satisfy the following properties:
\begin{subequations}\label{eq:p_function_properties} % A main label for the set of properties of p
\begin{align}
    &\begin{aligned}
        &\text{Condition 1: Normalization}\\
        &\sum_{k=1}^{N_p} p(s_{i,k},s_{i,-k}) = 1
     \end{aligned} \label{eq:p_normalization} \\
    &\begin{aligned}
        &\text{Condition 2: Monotonicity with respect to direct similarity scores}\\
        &p(s_{i,k_1},s_{i,-k_1}) \ge p(s_{i,k_2},s_{i,-k_2}) \Leftrightarrow s_{i,k_1} \ge s_{i,k_2}
     \end{aligned} \label{eq:loading_prob_dist_monotone}
\end{align}
\end{subequations}
\noindent The normalization condition ensures that the resulting $P$ is a probability distribution, while the monotonicity condition ensures that the assignment principle is adhered.

Overall, Equations \ref{eq:embedding_function}, \ref{eq:similarity_score}, and \ref{eq:loading_prob_dist} jointly defines the stochastic prompt generator (Equation \ref{eq:stochastic_generator_function}) utilized in our proposed approach. Beyond foundations in behavior theory, our strategy of constructing the persona loading function also addresses issues that arise with direct probability estimation of $P(\boldsymbol{Z_k}|\boldsymbol{d_i})$. Compared to directly establishing the relationship between $\boldsymbol{d_i}$ and $\boldsymbol{Z_k}$, our approach offers significantly improved robustness among two critical dimensions:
\begin{itemize}
    \item \textbf{Robustness to the number of personas}: A direct model would require an output of $|\mathcal{D}_k|$ dimensions. When $|\mathcal{D}_k|$ is large, a large number of parameters for $P(\boldsymbol{Z_k}|\boldsymbol{d_i})$ is needed, which significantly complicates computation and raises high data demand. Instead, our approach parameterize the matching function with embedding parameters, whose size can be compact and is invariant to $|\mathcal{D}_k|$, therefore mitigating the computation and data demand issues.
    \item \textbf{Robustness to data sparsity}: Sparsity of some population sub-groups due to survey coverage limitations is common in travel demand analysis, and this issue can make direct models highly susceptible to bias and variance due to the lack of observation on some $\boldsymbol{d}$s. Our approach mitigates this challenge by borrowing statistical strength from similar individuals from other population subgroups, thereby increasing the robustness of our approach against data sparsity.
\end{itemize}

\subsubsection{Estimating the persona loading function}

\begin{figure}[h!]
    \centering
    \includegraphics[width=1\linewidth]{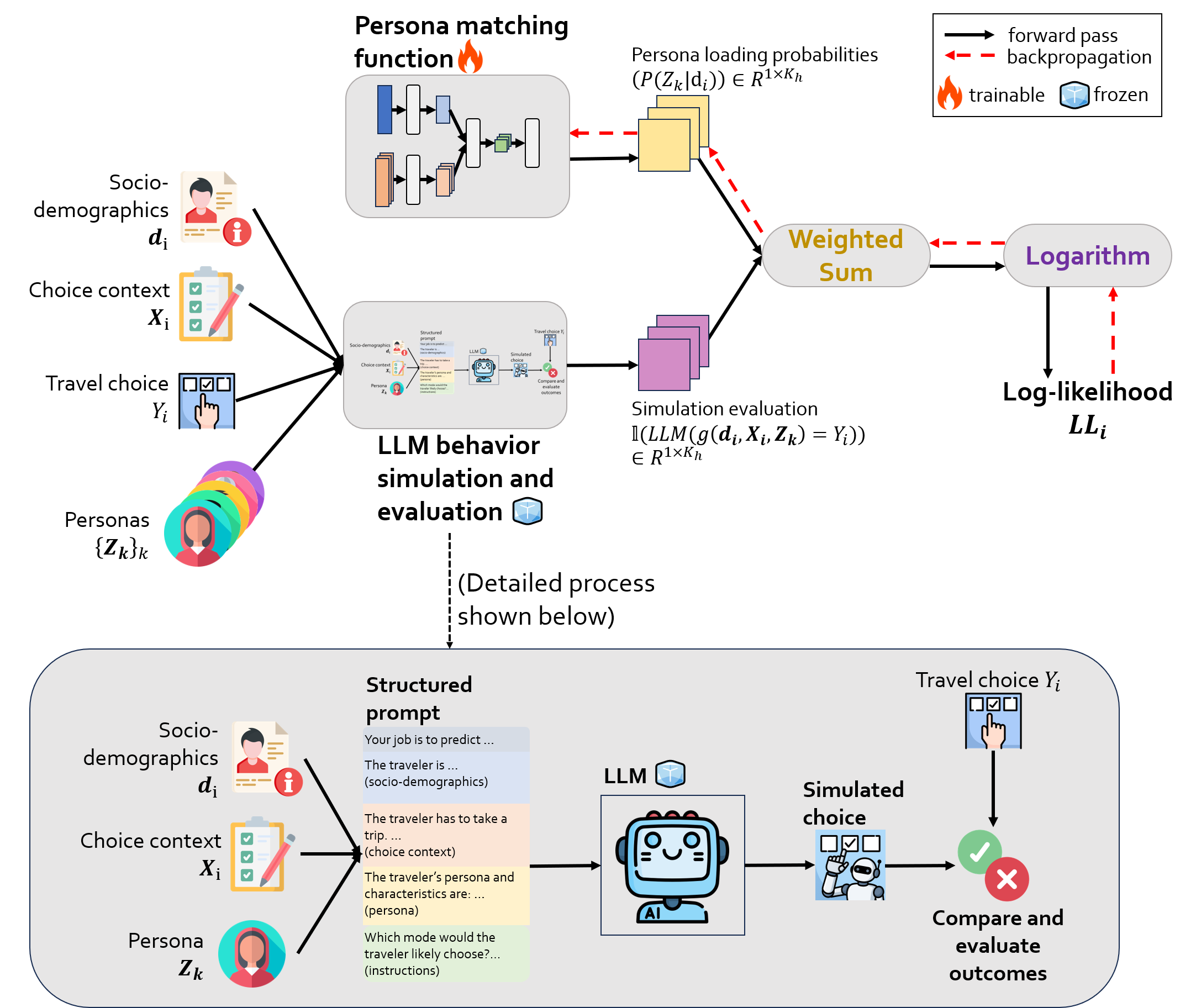}
    \caption{Illustration of the model estimation process}
    \label{fig:persona_training}
\end{figure}

With the definition of the persona loading function in \Cref{eq:loading_prob_dist}, we proceed to elaborate the model estimation process, which is illustrated in \Cref{fig:persona_training}.

The alignment problem (\ref{eq:population_level_alignment_problem_dataset}) can be transformed to a maximum likelihood estimation (MLE) problem, with the log-likelihood function $LL(\boldsymbol{\beta})$ defined as,
\begin{equation}
    LL(\boldsymbol{\beta})= \sum_{i\in \mathcal{D}_{\ell}}log(\sum_{k=1}^{K_h}P(\boldsymbol{Z_k}|\boldsymbol{d_i})\mathds{1}(LLM(g(\boldsymbol{d_i},\boldsymbol{X_i},\boldsymbol{Z_k}))=Y_i))
    \label{eq:likelihood_function}
\end{equation}

\noindent here, $\mathds{1}(LLM(g(\boldsymbol{d_i},\boldsymbol{X_i},\boldsymbol{Z_k}))=Y_i)$ is an indicator function that produces binary judgments on the correctness of LLM's behavior simulation results using the prompt $g(\boldsymbol{d_i},\boldsymbol{X_i},\boldsymbol{Z_k})$, and $P(\boldsymbol{Z_k}|\boldsymbol{d_i})$ signals the probability of the persona $\boldsymbol{Z_k}$ being loaded in the prompt generation process. Therefore,
\begin{equation*}
    \sum_{k=1}^{K_h}P(\boldsymbol{Z_k}|\boldsymbol{d_i})\mathds{1}(LLM(g(\boldsymbol{d_i},\boldsymbol{X_i},\boldsymbol{Z_k}))=Y_i)=\mathbb{E}(\mathds{1}(LLM(G_{\Phi}(\boldsymbol{d_i},\boldsymbol{X_i}))=Y_i))
\end{equation*}
\noindent represents the likelihood of individual observation $i\in \mathcal{D}_{\ell}$. Summarization of the individual log-likelihoods yields the log-likelihood function in \Cref{eq:likelihood_function}.

As a result, the MLE of the alignment problem can be formulated as:
\begin{equation}
    \hat{\boldsymbol{\beta}}_{MLE}=\arg max_{\boldsymbol{\beta}}LL(\boldsymbol{\beta})
    \label{eq:objective_function}
\end{equation}

%\begin{theorem}
%    \textbf{identifiability theorem here}
%    \texttt{Strategy now is to make assumption on L's property on s and discuss identifibility condition for some well-known s's and the rank of e. Let's try adding it within the week.}
%\end{theorem}

For the MLE problem in \Cref{eq:objective_function}, evaluation of the full likelihood function necessitates calling the LLM model to evaluate $\mathds{1}(LLM(g(\boldsymbol{d_i},\boldsymbol{X_i},\boldsymbol{Z_k}))=Y_i)$ for every persona in $\mathcal{D}_k$, which could be both time-consuming and costly. To reduce the LLM usage and the associated computational cost, we further utilize a Monte-Carlo stochastic EM algorithm to conduct the MLE estimation. An overview of the estimation procedure is shown in \Cref{alg:em_iterative_algorithm}.

\begin{algorithm}[h!]
\caption{Monte-Carlo Stochastic EM algorithm for MLE problem (\ref{eq:objective_function})}
\begin{algorithmic}[1]
\State Start: Initialize a set of embedding parameters $\boldsymbol{\beta}^{(0)}$. 
\State Set iteration counter $t=0$.
\State Set initial sample size $L=L_0$
\Repeat
    \State Increment iteration counter $t \leftarrow t+1$.
    \State Increment sample size $L\leftarrow L+1$
    \Statex \textbf{E-step:}
    \State Calculate the loading probabilities $P(\boldsymbol{Z_k}|\boldsymbol{d_i})$ for all $i=1,2,...,N_\ell$ and $k=1,2,...,K_h$ using $\boldsymbol{\beta}^{(t-1)}$ and \Cref{eq:loading_prob_dist}. 
    \State For every $i$, randomly sample $L$ personas by sampling from the calculated matching probability distribution $\{P(\boldsymbol{Z_k}|\boldsymbol{d_i})\}_k$. Denote these selected samples by index $l=1,2,...,L$. 
    \State Run LLM behavior simulation $LLM(g(\boldsymbol{d_{i}},\boldsymbol{X_{i}},\boldsymbol{Z_l}))$ to obtain simulated choice $\hat{Y}_{il}$ for each sampled persona $l$. 
    \State Compute the normalized weight $w_{il}$ for each sampled $(i,l)$ using $P(\boldsymbol{Z_l}|\boldsymbol{d_i})$ based on $\boldsymbol{\beta}^{(t-1)}$:
    \begin{equation*}
    w_{il}=
        \begin{cases}
        \frac{P(\boldsymbol{Z_l}|\boldsymbol{d_i}){\mathds{1}(\hat{Y}_{il}=Y_{i})}}{\sum_{j=1}^{L}(P(\boldsymbol{Z_j}|\boldsymbol{d_i})\mathds{1}(\hat{Y}_{ij}=Y_{i}))} & \sum_{j=1}^{L}\mathds{1}(\hat{Y}_{ij}=Y_{i}))>0\\
        0 &  \sum_{j=1}^{L}\mathds{1}(\hat{Y}_{ij}=Y_{i}))=0
        \end{cases}
    \end{equation*}
    \Statex \textbf{M-step:}
    \State Update the embedding parameter to  $\boldsymbol{\beta}^{(t)}$ by solving: 
    $$ \boldsymbol{\beta}^{(t)} = \arg\max_{\boldsymbol{\beta}}\sum_{i=1}^{N}\sum_{l=1}^{L}w_{il}\log(P(\boldsymbol{Z_l}|\boldsymbol{d_i})) $$ 
\Until{the parameters $\boldsymbol{\beta}^{(t)}$ converge.}
\State The final estimated parameters are $\hat{\boldsymbol{\beta}}_{MLE} = \boldsymbol{\beta}^{(t)}$.
\end{algorithmic}
\label{alg:em_iterative_algorithm}
\end{algorithm}

In the stochastic EM algorithm shown in \Cref{alg:em_iterative_algorithm}, the E-step use Monte-Carlo sampling to simulate the full-information log-likelihood in \Cref{eq:likelihood_function}. The purpose of the E-step is to provide a faithful estimation of the full-sample likelihood while reducing the number of LLM calls to reduce computational demand, as well as providing signals to adjust the embedding parameters $\boldsymbol{\beta}$ to improve the likelihood value. With sampling based off the prior $\boldsymbol{\beta}^{(t-1)}$, the simulated likelihood for sample $i$ can be expressed as:
\begin{equation}
    \Tilde{L}_i=\frac{\sum_{l=1}^L (P(\boldsymbol{Z_l}|\boldsymbol{d_i})\mathds{1}(\hat{Y}_{il}=Y_{i}))}{\sum_{l=1}^L P(\boldsymbol{Z_l}|\boldsymbol{d_i})}
    \label{eq:simulated_ind_likelihood}
\end{equation}

If the likelihood $\Tilde{L}_i$ is not zero for observation $i$ (i.e. at least one sampled persona leads to correct choice simulation), for each sampled persona $l=1,2,...,L$, its proportional contribution to the simulated likelihood in \Cref{eq:simulated_ind_likelihood} is:
\begin{equation}
    w_{il}=\frac{P(\boldsymbol{Z_l}|\boldsymbol{d_i}){\mathds{1}(\hat{Y}_{il}=Y_{i})}}{\sum_{j=1}^{L}(P(\boldsymbol{Z_j}|\boldsymbol{d_i})\mathds{1}(\hat{Y}_{ij}=Y_{i}))}
    \label{eq:contribute_portion}
\end{equation}
\noindent in which if $\hat{Y}_{ij}=Y_{i}$, then $w_{il}>0$ and the pair $i,l$ makes a positive contribution to the likelihood function. Otherwise, if $\hat{Y}_{ij}\neq Y_{i}$,  $w_{il}=0$ and the pair $i,l$ makes no contribution to the likelihood function.

Given the simulated results and the contributions of each pairings $w_{il}$, the M-step of \Cref{alg:em_iterative_algorithm} improves $\boldsymbol{\beta}$ by aligning the sampling probabilities with the normalized contributions $w_{il}$ of each match ${i,l}$:
\begin{equation}
    \max_{\boldsymbol{\beta}}\sum_{i=1}^{N}\sum_{l=1}^{L}w_{il}\log(P(\boldsymbol{Z_l}|\boldsymbol{d_i}))
\end{equation}
\noindent In this way, pairings that has no contributions to the likelihood function is devalued, and pairings with higher contributions to the likelihood function are rewarded. By iteratively conducting the E step and the M step, the likelihood function and the parameters $\boldsymbol{\beta}$ can be improved.

%\begin{theorem}
%    \textbf{local convergence theorem here}
%\end{theorem}

\section{Experiment} \label{sec:experiment}
Empirically, we test the performance of our proposed framework and compare it with established models in discrete choice modeling and LLM-based behavior simulation methods using a real-world survey dataset. The subsequent discussion in this sections elaborates on the core components of our empirical experiment.

\subsection{Dataset}
In our experiment, we use the publicly available Swissmetro survey dataset \citep{bierlaire2001acceptance}, which contains stated preference responses from 1,192 individuals in Switzerland, each providing nine responses to distinct choice contexts. The dataset features multiple choice records per respondent under varying conditions, enabling it to provide detailed observations specified in \Cref{subsec:prem_dataset}, and offers us the opportunity to infer the underlying personas based on travelers' stated preferences. The survey featured three transport modes: train, car, and Swissmetro (a proposed high-speed underground transit system). While most participants chose among all three modes, those without a car were presented with a choice between only the train and the Swissmetro. To establish a clear basis for evaluating and comparing performance while minimizing the impact of the choice set size, our empirical experiment centers on choice situations offering all three transport modes. After filtering out records with unknown information in the dataset, the processed dataset comprises 1,004 respondents and 9,036 choice records. An overview of variables used in the experiment and their descriptions is shown in \Cref{tab:var_sum}.

\begin{table}[h!]
	\caption{Summary of variables}\label{tab:var_sum}
	\begin{center}
		\begin{tabular}{p{0.3\linewidth}  p{0.55\linewidth}}
        \hline
  Variable & Description\\
  \hline
\multicolumn{2}{l}{\textit{Socio-demographic variables involved in embedding function}}\\

$\texttt{GENDER}$ & Dummy variable that equals 1 if the respondent is male, and zero otherwise.\\
$\texttt{AGE}$ & Dummy variable that equals 1 if the respondent's age is smaller than 25; equals 2 if between 25 and 39; equals 3 if between 40 and 54; equals 4 if between 55 and 65 and equals 5 otherwise.\\
$\texttt{INCOME}$ & Dummy variable that equals 1 if the respondent's annual income is under 50 thousand CHF; equals 2 if between 50 and 100 CHF; equals 3 otherwise.\\
$\texttt{GROUP}$ & Dummy variable that equals 1 if the respondent is in the car user group and equals 0 if the respondent is in the train user group.\\
\multicolumn{2}{l}{\textit{Alternative-invariant variables}}\\
$\texttt{PURPOSE}$ & Dummy variable that follows the leveling of: 1 for commuting, 2 for shopping, 3 for business, 4 for leisure, 5 for return from work, 6 for return from shopping, 7 for return from business, 8 for return from leisure.\\
$\texttt{FIRST}$ & Dummy variable for the class the respondent travels in that equals 1 if first-class and equals 0 otherwise.\\
$\texttt{WHO}$ & Dummy variable that signals who pays for the trip. Equals 1 if the trip is fully paid by the respondent; equals 2 if the payment is a half-half split by the respondent and the employer; and equals 3 if the payment is fully made by the employer.\\
$\texttt{LUGGAGE}$ &  Dummy variable related to the number of luggage items involved that equals 0 if there's no luggage, equals 1 if there's one piece of luggage, and equals 2 if there's multiple pieces of luggage.\\
$\texttt{GA}$ & Dummy variable related to the involvement of a rail system annual ticket that equals 1 if the traveler has one, and equals 0 otherwise.\\
\multicolumn{2}{l}{\textit{Alternative-specific variables}}\\
$\texttt{CO}$ & Price in Swiss Franc (CHF) of each alternative.\\
$\texttt{TT}$ &Travel time in minutes of each alternative.\\
$\texttt{HE}$ & Headway in minutes of each alternative. This variable is always 0 for the car.\\
\hline
		\end{tabular}
	\end{center}
\end{table}

From this filtered Swissmetro data, we constructed three distinct datasets for our experiments: the detailed observational dataset $\mathcal{D}_h$, the general observational dataset $\mathcal{D}_\ell$, and a testing dataset $\mathcal{D}_t$ for performance assessments. The dataset formulation procedure is described below:
\begin{enumerate}
    \item \textbf{The detailed observational dataset $\mathcal{D}_h$}: First, we randomly selected 250 respondents from the filtered data and extracted all their associated choice records (nine per respondent) to form $\mathcal{D}_h$. 
    \item \textbf{The general observational dataset $\mathcal{D}_\ell$}: Next, from the remainder of the dataset that is not included in $\mathcal{D}_h$, we randomly sampled 200 individual choice records to create $\mathcal{D}_\ell$. 
    \item \textbf{The testing dataset $\mathcal{D}_t$}: Finally, we randomly sample an additional 400 individual choice records from the remaining data after the formation of $\mathcal{D}_h$
  and $\mathcal{D}_\ell$ to constitute $\mathcal{D}_t$.
\end{enumerate}

The distributions of socio-demographic variables and choice outcomes in our three constructed datasets are shown in \Cref{tab:demo_sets}. Overall, while small differences in these distributions exist among the three datasets, it is clear that no significant shifts that substantially hamper the evaluation results exist between the datasets. Indeed, minor variations in sampling distributions are common in travel behavior research due to the inherent complexity and variance of data collection processes, and such variations in our experimental setting offer a valuable opportunity to examine the generalization ability under distribution shifts of our proposed approach.

\begin{table}[h!]
	\caption{Demographics and choice outcome of the datasets}\label{tab:demo_sets}
	\begin{center}
		\begin{tabular}{l l c c c}
  \hline
  Variable & Value & Detailed set & General set & Testing set\\
		    \hline
      \textit{Socio-demographics} & &\\
\texttt{GENDER} & Male & 79.6\% & 87.5\% & 85.0\%\\
& Female & 20.4\% & 12.5\% & 15.0\%\\
\texttt{AGE} & Below 25 & 2\% & 2.5\% & 3.5\%\\
& 25-39 & 29.2\% & 37.0\% & 29.7\%\\
& 40-54 & 42.4\% & 34.0\% & 39.3\%\\
& 55-65 & 22.0\% & 21.0\% & 20.5\%\\
& Above 65 & 4.4\% & 5.5\% & 7.0\%\\
\texttt{INCOME} & Under \$50k & 8.4\% & 16.5\% & 11.5\%\\
 & \$50k-\$100k & 40.0\% & 43.5\% & 39.5\%\\
  & Above \$100k & 51.6\% & 40.0\% & 49.0\%\\
\texttt{GROUP} & Train user & 25.2\% & 13.0\% & 21.5\%\\
& Car user & 74.8\% & 87.0\% & 78.2\%\\
\hline
\textit{Choices} & & &\\
Train & & 7.7\% & 8.0\% & 6.0\%\\
Swissmetro & & 59.4\% & 56.0\% & 53.3\%\\
Car & & 32.9\% & 36.0\% & 40.7\%\\
\hline
		\end{tabular}
	\end{center}
\end{table}\par

\subsection{Learning approach setup}

Throughout the experiment, we employ OpenAI's GPT-4o model via its API for all LLM calls. We select the GPT-4o model because it offers a strong balance between intelligence and operational cost-effectiveness (\$2.5/\$10 for 1 million input and output tokens, respectively). By utilizing the GPT-4o model, we aim to demonstrate the applicability of our approach beyond the advanced ``reasoning models" by showing our framework's efficacy with using the ``everyday task" LLM models.

In conducting the persona inference step specified in \Cref{eq:persona_inference}, we provide the LLM with each respondent's socio-demographics information as well as their nine context-choice pair in $\mathcal{D}_h$, and prompt the LLM to act as an expert to infer the respondent's persona. The design of these target personas is guided by established domain knowledge in choice modeling, particularly concerning mode choice behavior \citep{johansson2006effects,vij2013incorporating,paulssen2014values}. Specifically, we ask the LLM model to infer how much the respondent values the following factors: travel time, travel cost, flexibility, travel habit, comfort, and trip purpose. The LLM is prompted to give a numerical rate between 1 and 10 on the respondent's value, with increments in the rate indicating that the respondent values this factor more.

In using the LLM for behavior simulation, as shown in \Cref{eq:input_prompt_format}, for each sample in $\mathcal{D}_t$ we provide the LLM with the social-demographic information and the choice context of the sample, as well as the loaded persona. The LLM is then prompted to predict the individual's mode choice. We formulate the task for the LLM as a direct choice prediction problem rather than a role-playing problem based on recent findings that indicate this framing produces more robust and accurate simulations \citep{hansen2024simulating, anthis2025llm}. The LLM's textual prediction of the chosen mode is then programmatically extracted and transformed into a categorical variable for evaluation.

For the persona loading function, we use a linear embedding kernel based on socio-demographic variables,
\begin{equation}
    \boldsymbol{e_i}=(\boldsymbol{\beta}_{1}\cdot\texttt{GENDER}_i,\boldsymbol{\beta}_{2}\cdot\texttt{AGE}_i,\boldsymbol{\beta}_{3}\cdot\texttt{INCOME}_i,\boldsymbol{\beta_{4}}\cdot\texttt{GROUP}_i)
    \label{eq:embedding_used}
\end{equation}
\noindent in which $\boldsymbol{e_i}$ is a four-dimensional embedding vector ($D_e$=4). Each component of $e_i$ corresponds to the projection of a specific one-hot encoded socio-demographic variable in $\texttt{GENDER}_i$, $\texttt{AGE}_i$, $\texttt{INCOME}_i$, and $\texttt{GROUP}_i$, via its trainable parameter vectors $\boldsymbol{\beta_1}$ to $\boldsymbol{\beta_4}$. The dimensions of these parameter vectors match the number of categories in each one-hot encoded variable (2 for $\boldsymbol{\beta_1}$ and $\boldsymbol{\beta_4}$; 5 for $\boldsymbol{\beta_2}$; and 3 for $\boldsymbol{\beta_3}$).

For the embedding similarity measure function (\ref{eq:similarity_score}), we use the cosine similarity function:
\begin{equation}
    s_{i,k}=\frac{\boldsymbol{e_i}^T\boldsymbol{e_k}}{||\boldsymbol{e_i}||_2\cdot ||\boldsymbol{e_k}||_2}
    \label{eq:cosine_similarity_used}
\end{equation}
\noindent in which $||\boldsymbol{e_i}||_2$ and $||\boldsymbol{e_k}||_2$ are $L^2$ norms of vectors $\boldsymbol{e_i}$ and $\boldsymbol{e_k}$. The cosine similarity function satisfies the necessary conditions (\ref{eq:sim_non_negativity_bound}) to (\ref{eq:sim_self_similarity}) for the similarity measure.

For the persona loading function (\ref{eq:loading_prob_dist}), we use the weighted softmax function:
\begin{equation*}
    P(\boldsymbol{Z_k}|\boldsymbol{d_i})=\frac{e^{\lambda s_{i,k}}}{\sum_{j=1}^{K_h} e^{\lambda s_{i,j}}}
\end{equation*}
\noindent in which $\lambda$ is a scaling parameter, set to $40/3$ in our experiment. The weight softmax function adheres to the necessary conditions (\ref{eq:p_normalization}) and (\ref{eq:loading_prob_dist_monotone}) with respect to similarity scores.

In the training process, we utilize a generalized stochastic EM algorithm, which still follows the procedure shown in \Cref{alg:em_iterative_algorithm} but utilizes an adjusted weighting function at the end of the E step and an objective function with additional regularization for the M-step. The adjusted weighting function $\hat{w}$ is:
\begin{equation}
    \hat{w}_{il}=\frac{P(\boldsymbol{Z_l}|\boldsymbol{d_i}){\mathds{1}(\hat{Y}_{il}=Y_{i})}}{\sum_{j=1}^{L}(P(\boldsymbol{Z_j}|\boldsymbol{d_i})\mathds{1}(\hat{Y}_{ij}=Y_{i}))} \times \biggl(1- \alpha_e \times \mathds{1}(\prod_{j=1}^{L}\mathds{1}(\hat{Y}_{ij}=Y_{i})=1)\biggr)
    \label{eq:modified_weight}
\end{equation}
\noindent where $L$ is the number of personas sampled in the E-step, and $\alpha_e$ is a factor reducing the weight for those observations where all sampled personas lead to a correct prediction. We set $\alpha_e=0.5$ in our experiment, effectively reduce the weight of ``all-correct" observations by half. This down-weighting aims to prevent the model from overly focusing on ``easy" samples that offer uniformly positive signals, which could otherwise cause the learned embeddings to squeeze together in space, thereby reducing their discriminative power. Furthermore, the impact of this down-weighting naturally diminishes as $L$ increases in \Cref{alg:em_iterative_algorithm}.

Using the modified weights $\hat{w}_{il}$ in \Cref{eq:modified_weight}, in the M-step. In addition to the log-likelihood in \Cref{eq:objective_function}, we also add a regularization term $L_{reg}(\beta)$ to the M-step objective function,
\begin{equation}
    L_{reg}(\beta)=\sum_{m=1}^4(\frac{\sigma^2(\boldsymbol{\beta_m})}{\overline{\sigma^2}}-1)^2
    \label{eq:regularization_term}
\end{equation}
\noindent in which $\sigma^2(\boldsymbol{\beta_m})$ is the variance of the embedding parameters $\boldsymbol{\beta_m}$ in embedding dimension $m$, and $\overline{\sigma^2}$ is the mean of these variances across all embedding dimensions. This regularization term penalizes large disparities in the variances of the embedding dimensions. This is intended to prevent any single socio-demographic variable from dominating the cosine similarity in \Cref{eq:cosine_similarity_used}, which could occur if its embedding parameters become disproportionately large or varied values, leading to the cosine similarity measures being non-distinctive among other dimensions.

With the regularization term $L_{reg}$, we use the following regularization-informed objective function for the M-step in our model training process,
\begin{equation}
    \beta^{(t)} = \arg\max_{\beta}\sum_{i=1}^{N}\sum_{l=1}^{L}\hat{w}_{il}\log(P(\boldsymbol{Z_l}|\boldsymbol{d_i}))+\alpha_m\sum_{m=1}^4(\frac{\sigma^2(\boldsymbol{\beta_m})}{\overline{\sigma^2}}-1)^2
    \label{eq:objective_function_new}
\end{equation}
\noindent where $\alpha_m$ is the regularization strength, set to 0.4 in the experiment. This formulation, incorporating both the modified weights and the regularization, aims to address issues with the stochastic sampling and cosine similarity measure and promote a well-balanced latent space for generalization.

\subsection{Comparative models} \label{subsec:comp_models}

To evaluate the performance of our proposed framework, we compare it against representative established models in both traditional discrete choice modeling approaches and LLM-based behavioral simulation approaches:

\begin{itemize}
    \item \textbf{Multinomial logit (MNL) model}: The first model we use to compare is the MNL model. While not the most accurate, the MNL model is a classic baseline in DCM and a widely applied classic travel demand modeling and transportation planning practice. Therefore, we include it for comparison to assess our model. The MNL model's specification follows the benchmark specification of \cite{bierlaire2001acceptance}.

    \item \textbf{Zero-shot LLM}: The zero-shot LLM approach, which provides the LLM with only personal and context information in the prompt, is a widely applied approach in LLM-based behavior simulations \citep{tseng2024two,gao2024large}. In our case, the implementation of this approach involves providing the respondent's socio-demographic information and choice context solely of the to-be-predicted record to the LLM, which is then tasked to predict the respondent's choice directly from this information. The prompting structure for this benchmark follows \cite{liu2024can}.

    \item \textbf{Few-shot LLM}: As discussed in \Cref{subsec:LLM_alignment} few-shot prompting is another popular lightweight LLM alignment method for behavior simulation. This method augments the zero-shot contextual input with a pre-specified number of illustrative examples for the LLM to learn from. In our implementation, we first select the few-shot examples, comprising context-choice pairs, based on contextual similarity to the target record. The selection criteria follow the method outlined in \cite{liu2024can}. These examples and information about the target record are provided to the LLM, which is then tasked to infer the choice using both the target record's information and the patterns evident in the provided examples. We report the performance of this method using the optimal number of few-shot examples determined by preliminary tuning.

    \item \textbf{Same-group persona loading}: To compare against an alternative persona-based strategy, we implement an approach inspired by \cite{liu2024can}. In this benchmark, for a new observation, a persona $\boldsymbol{Z_k}$ is randomly selected from the subset of personas previously inferred for individuals belonging to the same socio-demographic group. This selected persona, along with the target individual's socio-demographics and choice context, is then used to prompt the LLM. This method contrasts with our proposed approach by utilizing a heuristic for persona selection rather than a learned persona loading function.
\end{itemize}

For all LLM-based benchmarks (zero-shot, few-shot, and same-group persona loading), to ensure fair comparisons, we consistently utilize the GPT-4o model identical to the LLM employed in our proposed framework via API access.

\subsection{Evaluation metrics}
We assess model performance from two primary perspectives. First, we evaluate the accuracy of the predicted aggregate choice shares, which is crucial for travel demand models. These forecasted choice shares directly influence predictions of traffic volume and overall transportation system performance, therefore carrying significant implications. Indeed, when comparing models with similar individual-level prediction accuracy, the one yielding more precise aggregate share predictions is generally preferred. We employ the Jensen–Shannon divergence (JSD) to quantify the differences between the predicted shares and the ground-truth shares for each alternative:
\begin{equation*}
    D_{JS}(q||p)=\frac{1}{2}\biggl(\sum_Yp(Y)log(\frac{p(Y)}{q(Y)})+\sum_Yq(Y)log(\frac{q(Y)}{p(Y)})\biggr)
\end{equation*}
\noindent in which $p(Y)$ and $q(Y)$ are the ground-truth share and predicted share of alternative $Y$. A lower $D_{JS}(q||p)$ signals that $p$ and $q$ are closer to each other, and only $p=q$ minimizes $D_{JS}(q||p)$ at 0.

Second, we assess the models' accuracy in predicting individual choices. For this, we utilize the macro F1-score and the weighted F1-score, defined as:
\begin{equation*}
\begin{aligned}
    &macro\:F_1=\frac{1}{|Y|}\sum_Y(2\times \frac{Precision_Y\times Recall_Y}{Precision_Y+Recall_Y})\\
    &weighted\:F_1=\sum_Y(\frac{n_Y}{N}\times 2\times \frac{Precision_Y\times Recall_Y}{Precision_Y+Recall_Y})\\
\end{aligned}
\end{equation*}
\noindent in which $N$ is the size of the testing dataset $\mathcal{D}_t$, $n_Y$ is the number of records in $\mathcal{D}_t$ that are predicted to choose alternative $Y$, $|Y|$ is the number of alternatives, and $Precision_Y$ and $Recall_Y$ are the precision and recall rate of an alternative. The macro F1-score evaluates a model's average performance by treating each class as equally important, aiming for balanced efficacy across all alternatives. In contrast, the weighted F1-score assesses overall performance by weighting each class's contribution by its prevalence, thereby prioritizing effectiveness on more frequently chosen alternatives. In both cases, a higher $F_1$ score signals higher prediction accuracy on the individual level, while only a perfect predictor would yield $F_1=1$.

\section{Results} \label{sec:results}
In this section, we present the results of our empirical experiment. Our experimental insights are twofold: first, the results showcase the performance of our proposed method and its comparison with established methods, which are presented in \Cref{subsec:res_perf_comp}. Second, the training embedding parameters can be interpreted to reveal underlying behavioral similarities between socio-demographic groups, which are illustrated in \Cref{subsec:res_embed_interp}.

\subsection{Performance comparison} \label{subsec:res_perf_comp}

The results of our proposed model, along with its comparison to the models in \Cref{subsec:comp_models}, are presented in \Cref{tab:pred_results}.

\begin{table}[h!]
	\caption{Comparison of our model with established models on prediction performance}\label{tab:pred_results}
	\begin{center}
		\begin{tabular}{l |  c c c | c c c}
  \hline
  \multirow{2}{*}{Method}  & \multicolumn{3}{c|}{Mode split} &  Jensen-Shannon & Macro & Weighted\\
  & Train & Swissmetro & Car & Divergence & F1-score & F1-score\\
  \cline{3-5}
		    \hline
Ground truth  & 6.0\% & 53.3\% & 40.7\% & 0.000 & 1.000 & 1.000\\
\hline
MNL  & 2\% & 77.5\% & 20.5\% & 0.483 & 0.474 & 0.606\\
Zero-shot LLM  & 3.3\% & 70.0\% & 26.7\% & 0.216 & 0.407  & 0.543\\
Few-shot LLM   & 3.5\% & 65\% & 31.5\% & 0.108 & 0.429 & 0.594\\
Same-group persona & 3.5\% & 49.7\% & 46.8\% & 0.044 & 0.542 & 0.657\\
Our method & 4.0\% & 51.7\% & 44.3\% &\textbf{0.021} & \textbf{0.556} & \textbf{0.683}\\
\hline
		\end{tabular}
	\end{center}
\end{table}\par

The results summarized in \Cref{tab:pred_results} demonstrate the strong performance of our proposed method across both aggregate and individual prediction levels. Our approach achieves the lowest JSD from the ground truth mode shares among all models tested, and concurrently, it records the highest macro F1-score and weighted F1-score for individual mode choice prediction. Compared to the MNL model, our proposed method reduces the JSD between the predicted mode share and the ground truth mode share by 95.7\% (from 0.483 to 0.021) and increases the F1-score by 17.3\% and 12.7\% on the macro level (from 0.393 to 0.556) and the weighted level (from 0.527 to 0.683), correspondingly. Compared to the lightweight few-shot LLM method, our method decreases the JSD by 90.3\% (from 0.216 to 0.021) and enhances the prediction F1-scores by 29.6\% (macro level, from 0.429 to 0.556) and 15.0\% (weighted level, from 0.543 to 0.683). Overall, the results clearly indicate that our proposed method not only generates more accurate simulations of individual travel choice behavior than existing methods but also provides more precise estimations of aggregate mode shares.

\begin{figure}[h!]
    \centering
    \includegraphics[width=0.8\linewidth]{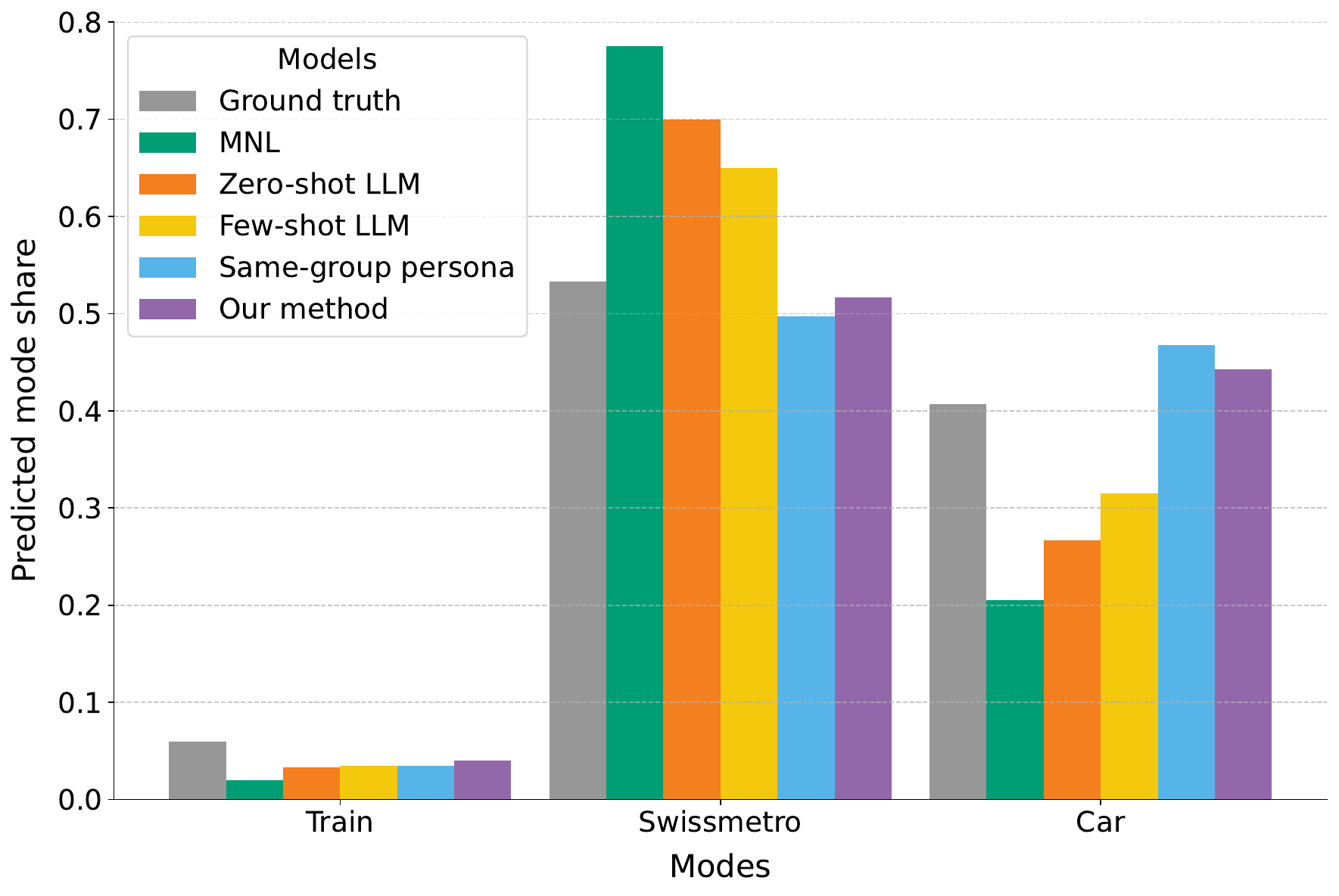}
    \caption{Comparison of predicted mode shares}
    \label{fig:mode_share_comp}
\end{figure}

To comprehensively assess our model's predictive performance, the subsequent discussion examines in detail the results for both aggregate mode share prediction and individual choice prediction. First, on the aggregate level, comparisons between the predicted mode shares of each model is shown in \Cref{fig:mode_share_comp}.

The detailed mode share comparisons presented in \Cref{fig:mode_share_comp} reveal that our method's predictions most closely align with the ground truth shares across all three transport modes. For the train alternative, which is the least-frequently selected by the respondents, all models tend to underestimate its mode share. The MNL model exhibits the largest underestimation, while our proposed method shows the least. The other LLM-based benchmarks produce similar underestimations for this mode. 
Regarding the Swissmetro mode, the MNL model overestimates its share while underestimating that of the car. The zero-shot LLM demonstrates a strong preference for Swissmetro, overestimating its share by 31.3\% and underestimating the share of car by 34.3\%. A possible explanation of this is that LLMs may have their own tendencies and values regarding different travel characteristics in travel choice prediction when the persona is not provided to them. The GPT-4o model may value travel time the most, which is consistent with the finding of our previous work \citep{liu2024can} on its predecessor, GPT-4, and therefore having biased economic behavior than human travelers. While the few-shot LLM method partially mitigates this tendency (reducing Swissmetro's predicted share by 5\% compared to zero-shot), it does not fully correct the issue. The same-group persona method, on the other hand, overestimates the usage of cars and underestimates the Swissmetro's share, reflecting that solely sampling personas from the same group may also create biases in behavioral representation. In conclusion, the superior aggregate performance of our proposed method highlights its ability to effectively mitigate these varied model-specific biases and achieve more accurate mode share predictions.

\begin{figure}[htbp]
\centering
    \subfloat[][MNL]{\includegraphics[width=0.45\linewidth]{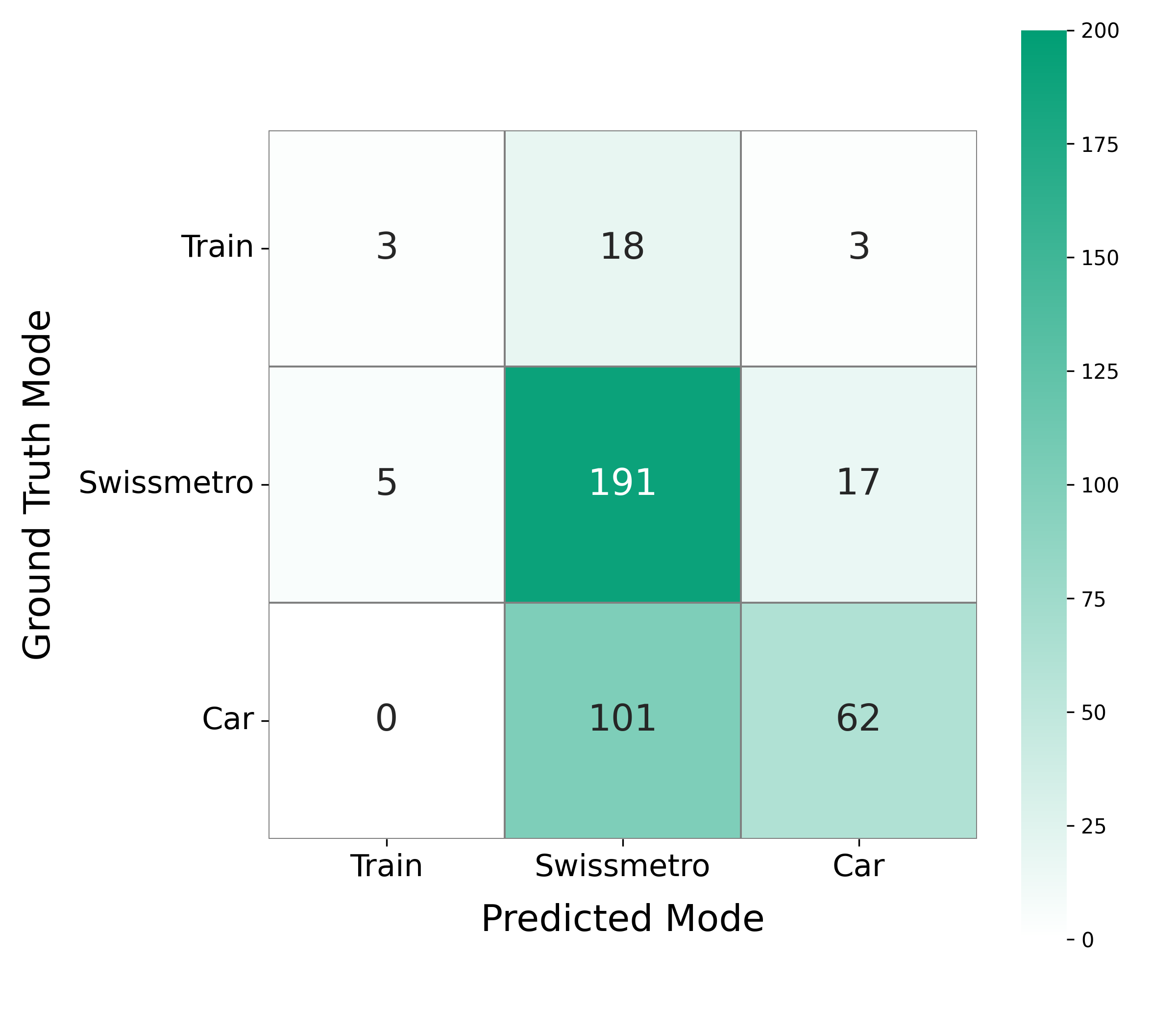}}
	\subfloat[][Zero-shot LLM]{\includegraphics[width=0.45\linewidth]{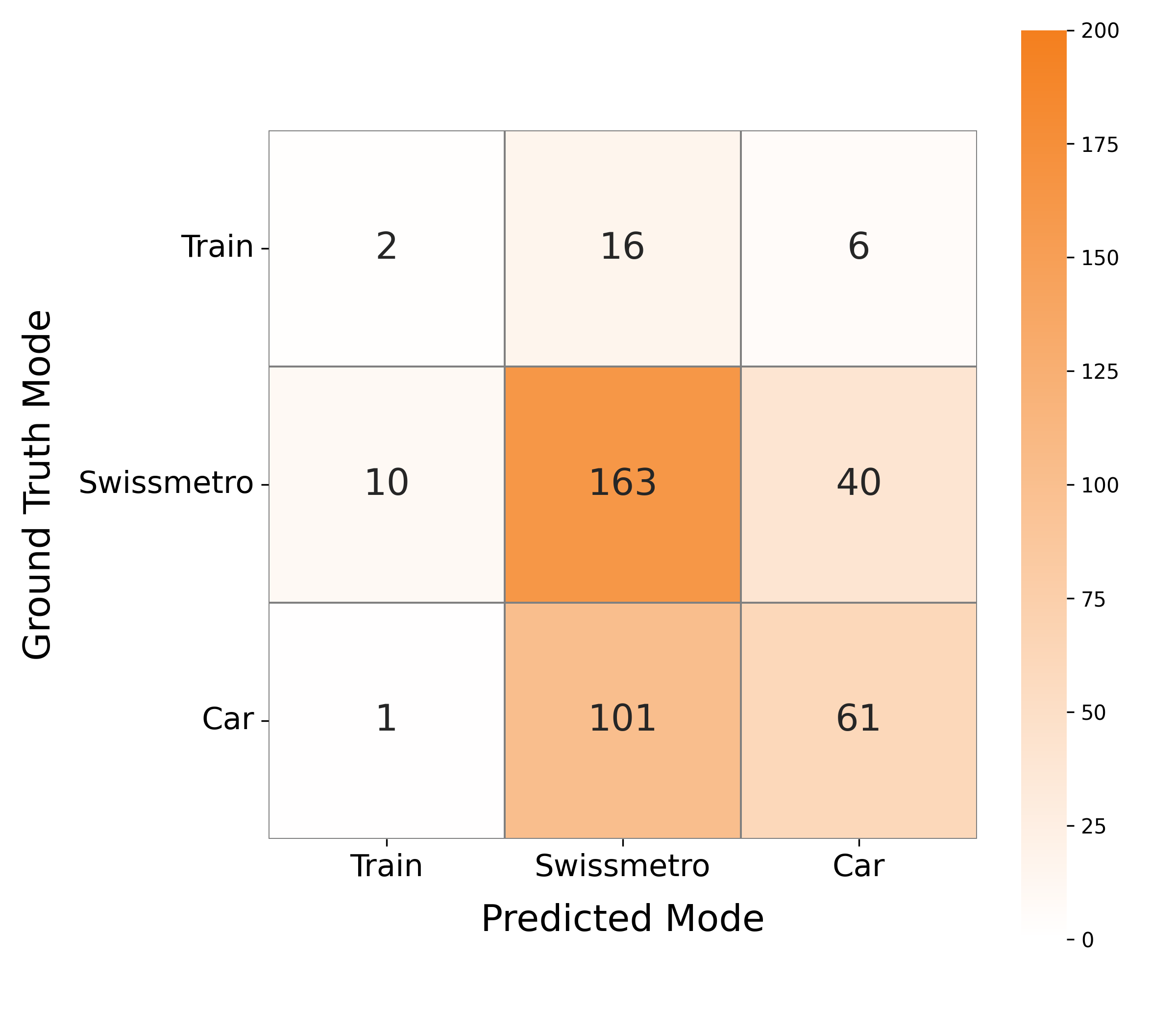}}
    \newline
	\subfloat[][Few-shot LLM]{\includegraphics[width=0.45\linewidth]{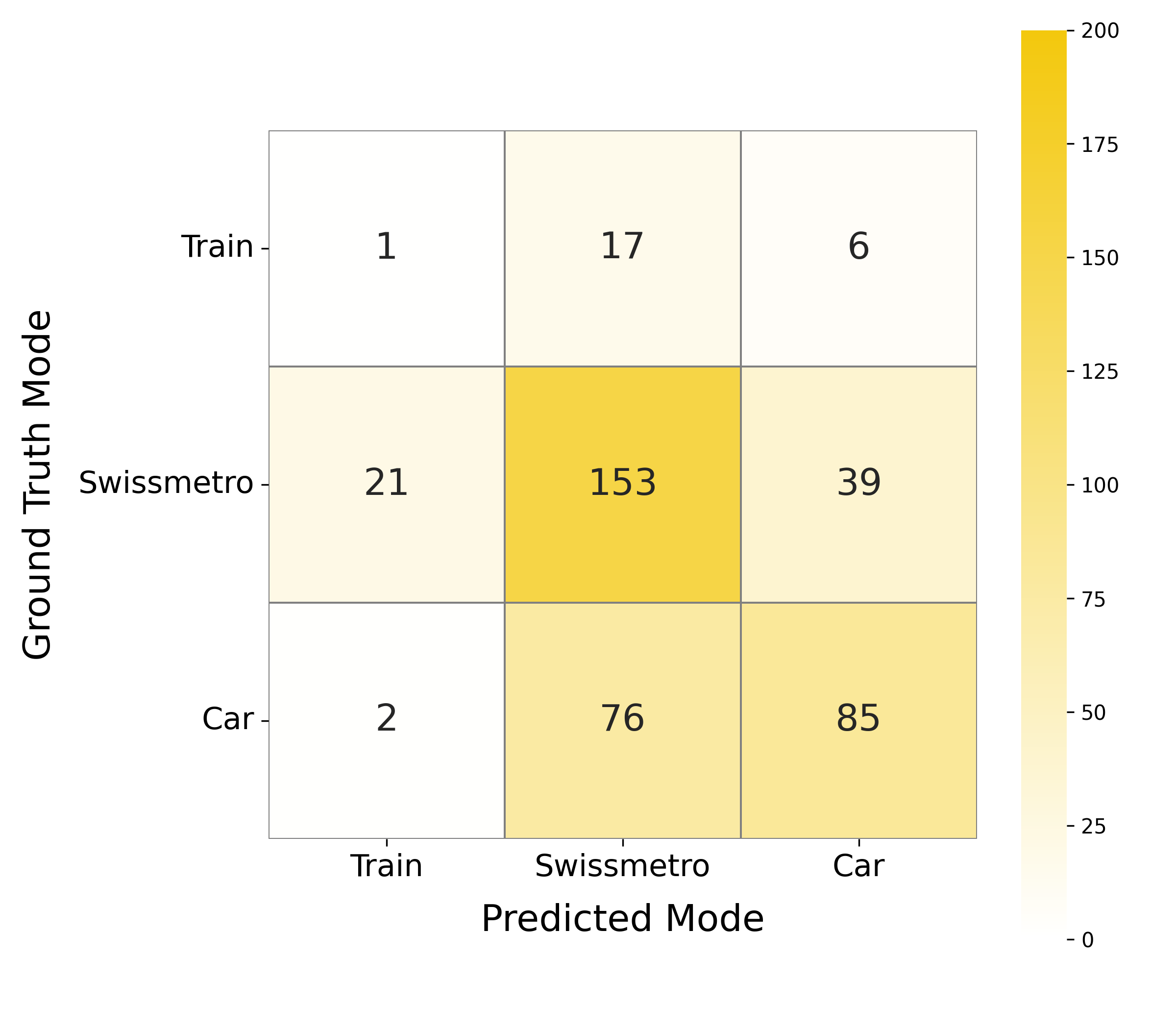}}
	\subfloat[][Same-group persona]{\includegraphics[width=0.45\linewidth]{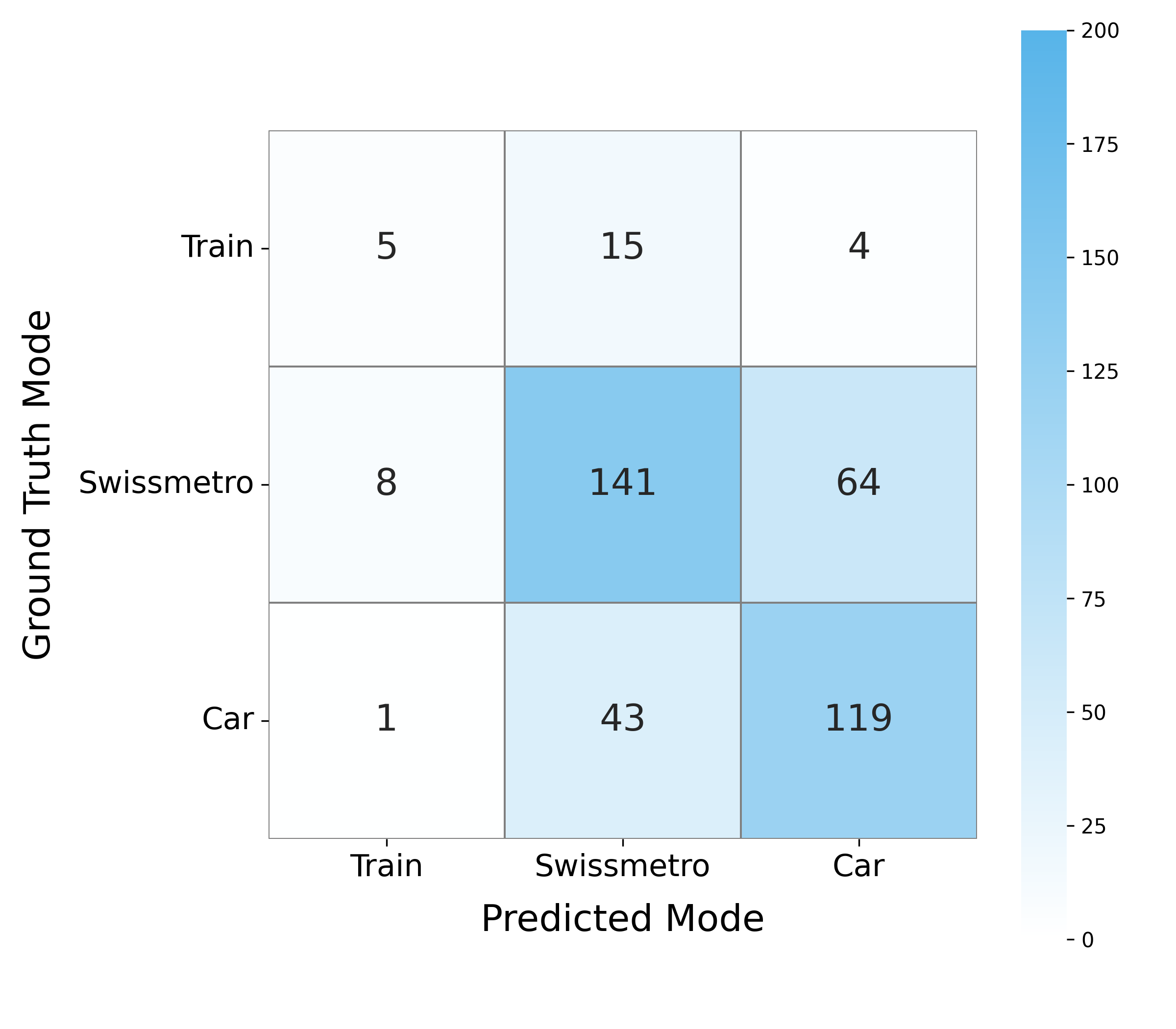}}
	\newline
	\subfloat[][Our method]{\includegraphics[width=0.45\linewidth]{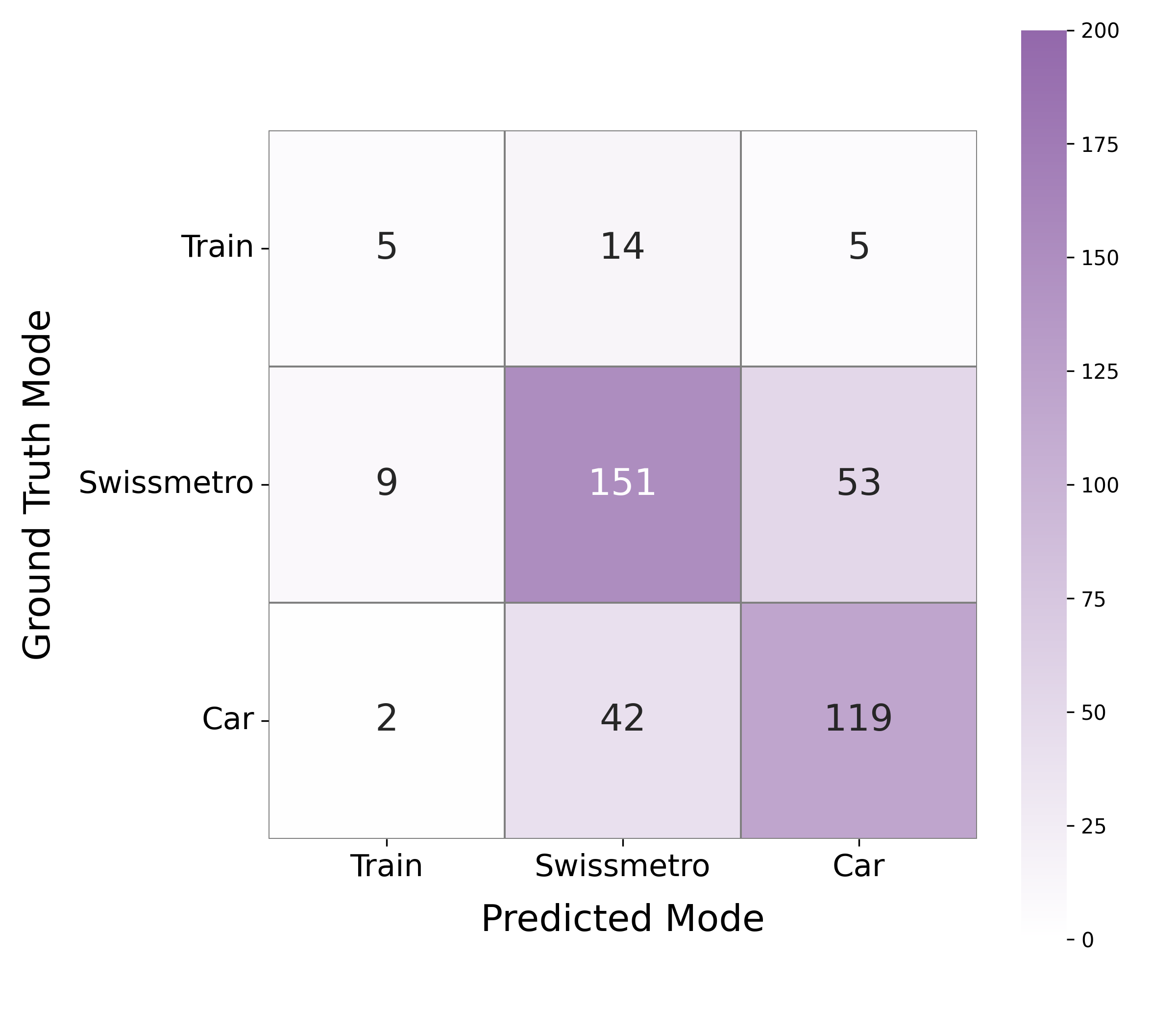}}
	\caption{Comparison of confusion matrices for all models}
	\label{fig:confusion_matrices}
\end{figure}

Next, on the individual behavior prediction level, we report the confusion matrices of all used models in \Cref{fig:confusion_matrices} to gain a detailed look.

The confusion matrices in \Cref{fig:confusion_matrices} detail the models' prediction accuracy on individual choice behavior among each mode. Overall, the MNL model, for example, is most accurate for predicting Swissmetro choices but exhibits substantially lower accuracy for individuals choosing car or train. The zero-shot LLM displays similar deficiencies in predicting non-Swissmetro choices, and the few-shot LLM model gains an increment in prediction accuracy by significantly improving the car predictions at the small expense of reduced accuracy on Swissmetro choosers. The introduction of personas yields further gains, as both persona-based methods significantly improve prediction accuracy for car and train choices relative to the non-persona LLM approaches. Our method further stands out by enhancing predictive accuracy for the Swissmetro mode without diminishing performance for the other two modes. This results in a superior overall balance, leading to the most accurate individual-level choice predictions among all evaluated models.

\subsection{Embedding interpretation} \label{subsec:res_embed_interp}

Beyond good prediction performance, our framework is also interpretable as the embedding vectors in \Cref{eq:embedding_used} are directly connected with socio-demographic variables, and their relative position in space indicates the behavior similarity between the socio-demographic groups they stand for. Therefore, by examining the learned embedding parameters, we can also obtain insights into the connection between socio-demographics and economic preferences in the population. Our following discussion showcases the interpretation and takeaways from the embedding parameters.

\begin{figure}[h!]
    \centering
    \includegraphics[width=0.55\linewidth]{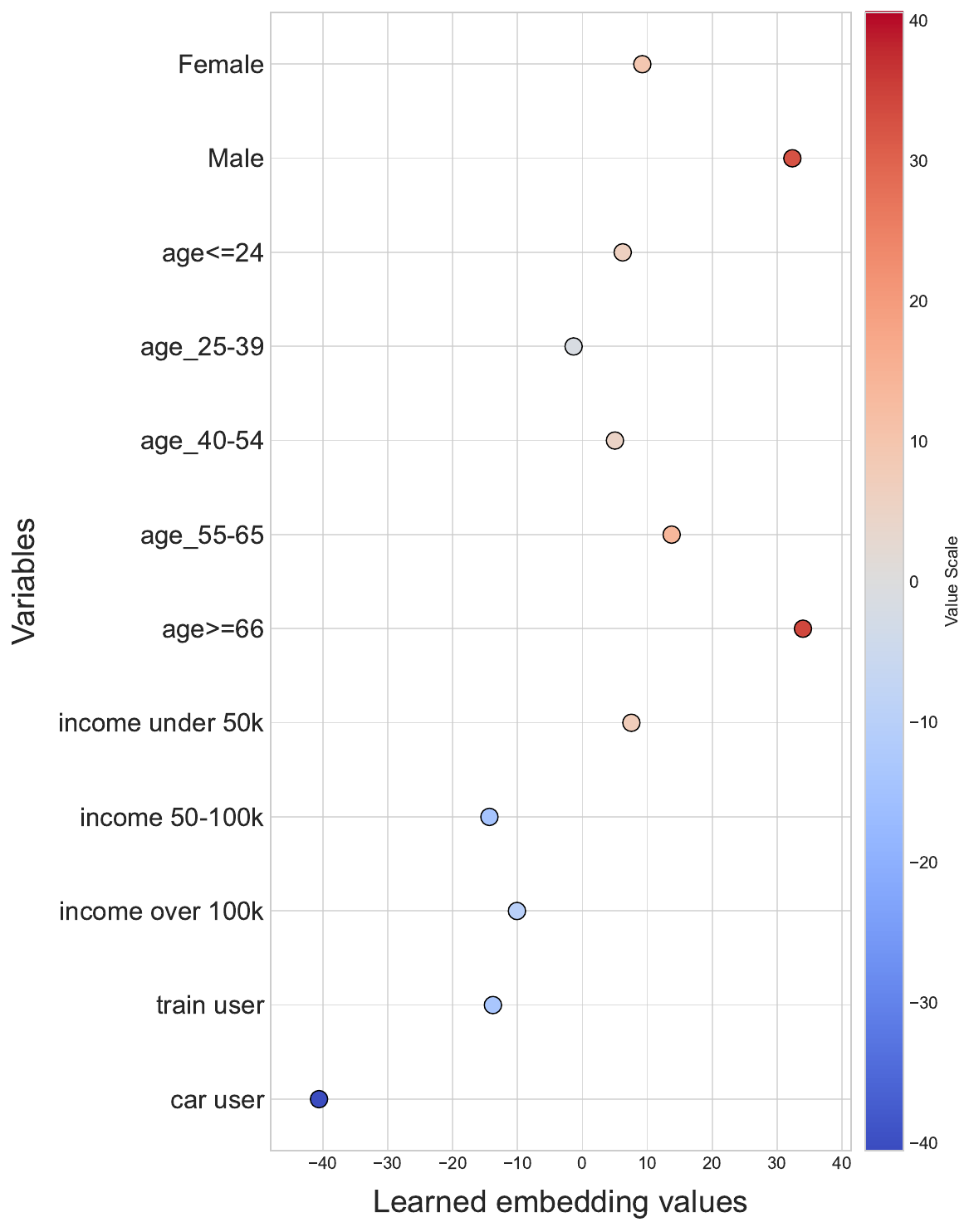}
    \caption{Values of learned embedding parameters }
    \label{fig:embed_param_values}
\end{figure}

First, we present a visualization of the values of the fitted embedding parameters $\boldsymbol{\beta_1}$ to $\boldsymbol{\beta_4}$ in \Cref{fig:embed_param_values}, which signal the positional proximity of the embedding projections in their respective dimensions.

The learned embedding values in \Cref{fig:embed_param_values} offer valuable behavioral insights. Firstly, the absolute magnitudes of these parameters reveal that certain socio-demographic characteristics are projected to have more distinctive embedded positions. For instance, categories such as ``male", ``age above 65", and ``car user" have notably large parameter values. This indicates that the model has learned to strongly differentiate these groups from others within the same socio-demographic variable, signaling their distinct behavioral relevance in the context of persona loading. Second, the parameter values across the categories of each socio-demographic variable also present rich insights. For example, on the age dimension, the embedding value of ``age 25-34" is the sole negative one, and furthermore, among the positive values, ``age above 66" is clearly distinct from the other three, which are positioned more closely. This indicates that elders and some young adults have different personas from the other age groups. Regarding income, lower income groups are projected to the positive axis, while middle and high income groups are projected close together on the negative axis, signaling a clear distinction in their underlying economic behavior.

\begin{figure}[h!]
    \centering
    \includegraphics[width=0.8\linewidth]{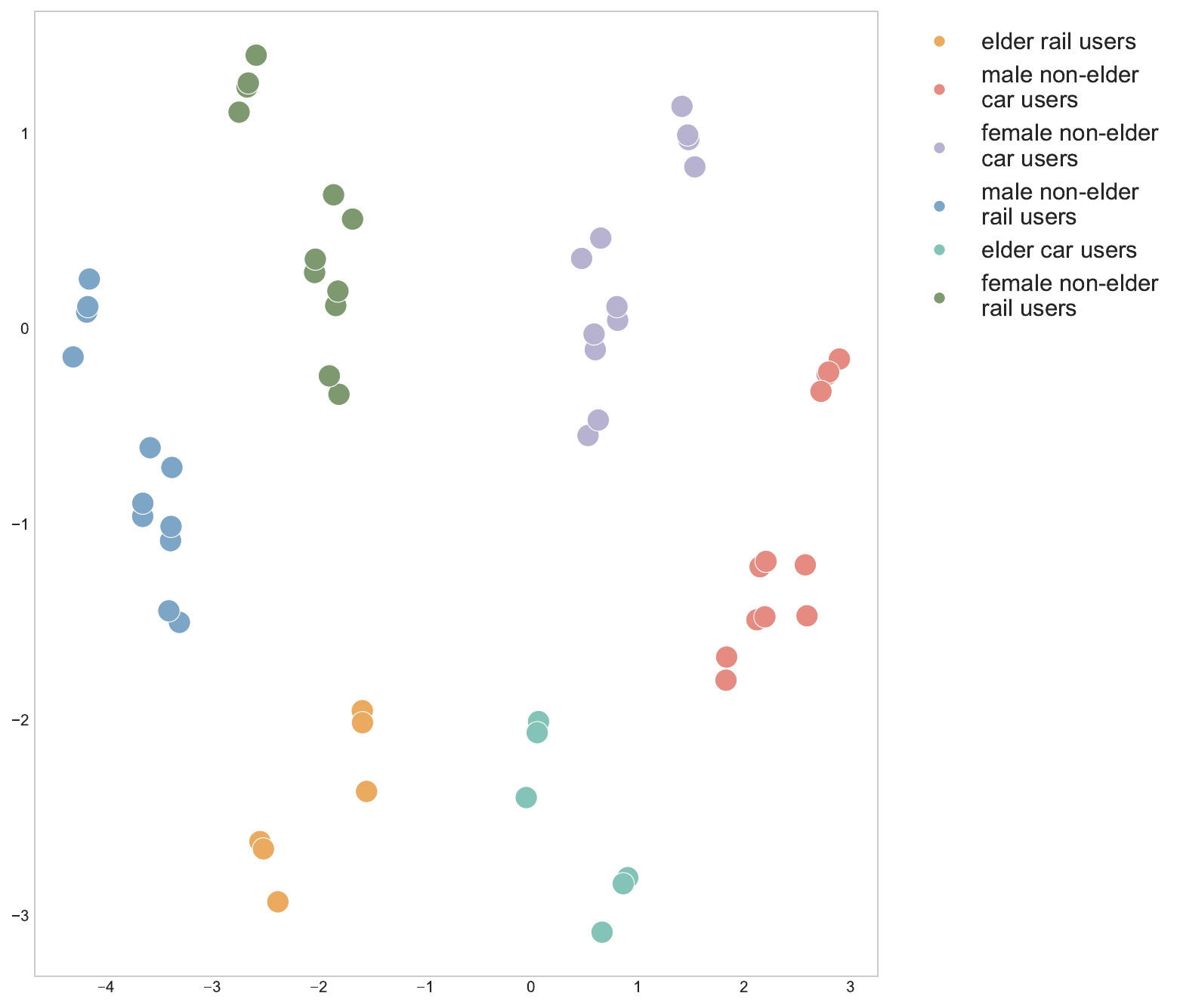}
    \caption{Clusters of socio-demographic groups in the embedding space}
    \label{fig:embed_groups}
\end{figure}

As the proximity of the embedded vectors signals behavioral similarity, we can also discover latent groups of different socio-economics that have similar behavior. For our experiment, we present the discovered behavior clusters based off the training embeddings in \Cref{fig:embed_groups}. The clusters are formed using the K-means clustering algorithm, and the projection to 2-D space in the illustration is conducted by the t-distributed stochastic neighbor embedding algorithm.

Overall, \Cref{fig:embed_groups} reveals six distinct behavior clusters in the population. Elder persons have two distinct groups based on their car/rail usage, while other people are split into four groups based on gender and membership in the car/rail group. The economic behavior of those within the same group are more similar, while that of those across groups is less similar. These revealed cluster shows that old age, past mode usage, and gender significantly affect the underlying economic preferences of individuals in the Swissmetro mode choice case.

\section{Conclusion} \label{sec:conclusion}
In this paper, we propose a novel alignment method to enable LLMs to simulate human travel choice behavior effectively. Our approach conditions an LLM using a structured prompt that incorporates not only socio-demographic information and the choice context but also a dynamically loaded persona describing the traveler's preferences and behavioral characteristics. Central to the alignment process is a learnable persona loading function, which utilizes representations of socio-demographics to map individuals to suitable personas from a basis set inferred from observational data. This persona loading function is estimated through a Monte-Carlo stochastic EM algorithm. Our empirical evaluation on the Swissmetro dataset demonstrated that our proposed framework significantly outperforms baseline DCMs as well as established lightweight LLM alignments methods in both aggregate choice share prediction and individual choice accuracy. Furthermore, the interpretability afforded by the learned parameters within our persona loading mechanism offers rich behavior insight, highlighting factors such as old age, past mode usage, and gender as significant impact factors on travelers' economic behavior in this case study. The proposed methodology offers significant implications for travel demand modeling by offering a pathway to developing more adaptable LLM-based behavior modeling tools. By enabling the effective use of readily available, pre-trained LLMs without the necessity for full supervised fine-tuning, our approach also enhances the accessibility of accurate LLM-based behavioral simulation for researchers and practitioners.

Future research can focus on several promising directions to enhance LLM's ability in behavior simulation. One possible direction is enhancing prompting. While our structured and persona-enhanced prompting approach works well in the Swissmetro case, further experimental insights for prompting strategies for a wider array of travel behavior scenarios are still lacking. Future work could thus explore further refinements to prompting mechanisms to establish a stronger foundation for LLM alignment, alongside investigating the potential of generalizable meta-prompts tailored for travel demand modeling. Another direction is examining model generalizability with multi-context data. Although our proposed framework is general and could allow for training data across different choice contexts, its empirical evaluation in this study was necessarily focused on the Swissmetro case due to data access constraints. Future work could therefore test and refine our framework within data fusion contexts, offering further insights into its cross-contextual robustness and informing directions for further enhancements and wider applications.

\newpage
\bibliographystyle{apalike}
%\nocite{*}
%\bibliography{mybib}

% \clearpage
% \begin{appendices}
% \input{appendix.tex}
% \end{appendices}
\end{document}